\documentclass[lettersize,journal]{IEEEtran}
\usepackage{amsmath,amsfonts}
\usepackage{algorithmic}
\usepackage{algorithm}
\usepackage{array}
\usepackage[caption=true,font=footnotesize,labelfont=sf,textfont=sf]{subfig}
\usepackage{textcomp}
\usepackage{stfloats}
\usepackage{url}
\usepackage{verbatim}
\usepackage{graphicx}
\usepackage[nocompress,noadjust]{cite}

\usepackage{amssymb}
\usepackage{hyperref}
\usepackage{tabularx} 
\usepackage{booktabs} 
\usepackage{ragged2e} 
\usepackage{array} 
\usepackage{float}
\usepackage{multirow} 
\usepackage{CJKutf8}
\usepackage{balance}
\usepackage{threeparttable}

\begin{document}

\title{A GAN and LLM-Driven Data Augmentation Framework for Dynamic Linguistic Pattern Modeling in Chinese Sarcasm Detection}

\author{Wenxian Wang, Xiaohu Luo, Junfeng Hao, Xiaoming Gu, Xingshu Chen, Zhu Wang, \\and Haizhou Wang*,~\IEEEmembership{Member,~IEEE,}
        % <-this % stops a space
\thanks{This work was supported in part by the National Natural Science Foundation of China (NSFC) under Grant 62572331.\textit{(*Corresponding author: Haizhou Wang.)}}% <-this % stops a space
\thanks{Wenxian Wang and Xingshu Chen are with the Key Laboratory of Data Protection and Intelligent Management, the Cyber Science Research Institute, Sichuan University, Chengdu 610207, China (e-mail: catean@scu.edu.cn; chenxsh@scu.edu.cn).}
\thanks{Xiaohu Luo, Junfeng Hao and Haizhou Wang are with the School of Cyber Science and Engineering, Sichuan University, Chengdu 610207, China (e-mail: xiaohuluo@stu.scu.edu.cn; haojunfeng@stu.scu.edu.cn; whzh.nc@scu.edu.cn).}
\thanks{Xiaoming Gu is with the State Key Laboratory of Fluid Power and Mechatronic Systems, Zhejiang University, Hangzhou 310027, China (e-mail: xiaominggu@zju.edu.cn).}
\thanks{Zhu Wang is with the Law School, Sichuan University, Chengdu 610207, China (e-mail: wangzhu@scu.edu.cn).}}

\maketitle

\begin{abstract}
Sarcasm is a rhetorical device that expresses criticism or emphasizes characteristics of certain individuals or situations through exaggeration, irony, or comparison. Existing methods for Chinese sarcasm detection are constrained by limited datasets and high construction costs. And they mainly focus on the textual features, overlooking user-specific linguistic patterns that shape how opinions and emotions are expressed. This paper proposes a Generative Adversarial Network (GAN) and Large Language Model (LLM)-driven data augmentation framework to dynamically model users’ linguistic patterns for enhanced Chinese sarcasm detection. First, we collect raw data from various topics on Sina Weibo. Then, we train a GAN on these data and apply a GPT-3.5 based data augmentation technique to synthesize an extended sarcastic comment dataset, named SinaSarc. This dataset contains target comments, contextual information, and user historical behavior. Finally, we extend the BERT architecture to incorporate multi-dimensional information, particularly user historical behavior, enabling the model to capture dynamic linguistic patterns and uncover implicit sarcastic cues in comments. Experimental results demonstrate the effectiveness of our proposed method. Specifically, our model achieves the highest F1-scores on both the non-sarcastic and sarcastic categories, with values of 0.9138 and 0.9151 respectively, which outperforms all existing state-of-the-art (SOTA) approaches. This study presents a novel framework for dynamically modeling users’ long-term linguistic patterns in Chinese sarcasm detection, contributing to both dataset construction and methodological advancement in this field.
\end{abstract}


\section{Introduction}
\begin{figure}[H]
	\centering
	\includegraphics[width=\columnwidth]{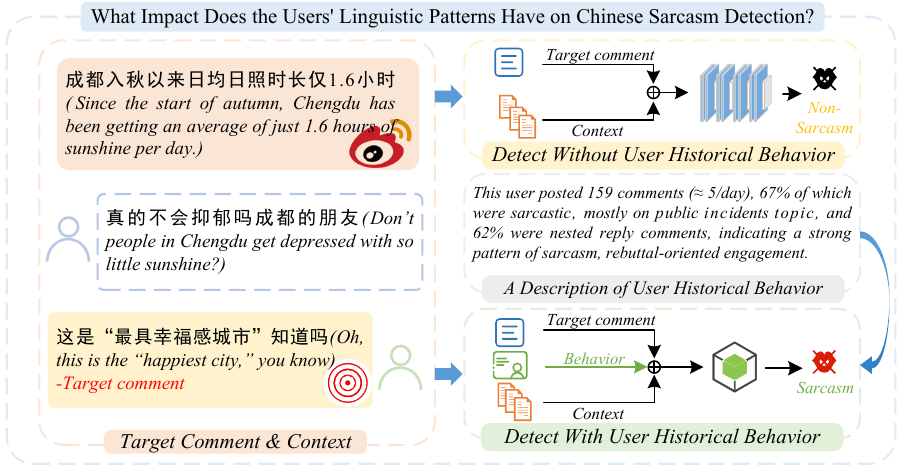}
	\caption{Here is an example demonstrating the importance of users’ long-term linguistic patterns for sarcasm detection. By learning such patterns from user historical behavior, our model can more accurately identify sarcastic comments.}
	\label{fig0}
\end{figure}

\IEEEPARstart{S}{arcasm} is a rhetorical device that uses subtle language to criticize some viewpoints, individuals, or events. In recent years, sarcasm has become increasingly prevalent on social media, as it enables users to convey intense emotions in a more vivid manner. However, the subtlety of sarcasm poses a significant challenge to sarcasm detection methods. This is because the sarcastic nature of a comment is closely tied to the user’s linguistic patterns; the same sentence can carry completely different meanings when spoken by different people. For instance, as illustrated in Fig.~\ref{fig0}, in response to the social issue of insufficient daily sunshine in Chengdu, a user comments by quoting the official promotional slogan “the happiest city.” Based solely on the immediate context, the comment could be interpreted either as a positive endorsement or as sarcastic comment. However, by learning the user's long-term linguistic pattern from his historical behavior, especially the strong tendency toward sarcastic expression, our model can more accurately identify the comment as sarcastic.
\IEEEpubidadjcol
	
\subsection{Background}
Existing sarcasm detection methods mainly fall into three categories: rule-based approaches, machine learning
approaches, and deep learning approaches. Rule-based approaches identify sarcastic features via manually defined linguistic or emotional rules \cite{ref4, ref6, ref17, ref18}; some studies using features like emoji polarity ~\cite{ref86} have achieved high accuracy for known sarcasm patterns, but they struggle with the implicit and context-dependent nature of sarcasm due to inability to capture subtle linguistic cues. This limitation led to the shift to machine learning approaches, which improve performance by learning sarcasm’s statistical characteristics from data ~\cite{ref15, ref23, ref25, ref27, ref87, ref100}. But these methods still rely heavily on handcrafted features, restricting scalability and robustness. Deep learning approaches thus emerged as a more effective alternative: they automatically capture deep semantic and contextual features through complex neural networks ~\cite{ref7, ref85, ref90, ref95, ref96, ref97, ref98, ref99}, such as dual-module BiLSTM ~\cite{ref29}, multimodal frameworks with quantum probability ~\cite{ref88}, knowledge-enhanced graph networks ~\cite{ref89}. These methods reduce reliance on manual feature design and achieve substantial success in handling sarcasm’s complex contextual nature.

\subsection{Challenges}
Although substantial research has been devoted to sarcasm detection in English, comparatively limited attention has been paid to the Chinese context. Currently, Chinese sarcasm detection research faces two key challenges.
\begin{enumerate}
	\item \textbf{Existing methods overlook the role of users' long-term linguistic patterns in sarcasm detection.} Sarcasm is a subtle form of expression that is closely tied to users’ long-term linguistic patterns. Due to individual differences in expressive style, the same words may convey different meanings when produced by different users. Therefore, it is necessary to develop methods that can capture and model such long-term linguistic patterns for more accurate sarcasm detection.
	\item \textbf{Lack of large-scale annotated Chinese sarcasm datasets with user historical behavior.} Existing Chinese sarcasm datasets are scarce and costly to construct. Moreover, most datasets focus on target texts and their immediate context, while overlooking user-level information derived from their historical behavior. Such information reflects users’ long-term linguistic patterns and behavioral tendencies, which provide implicit cues for sarcasm detection.
\end{enumerate}

\subsection{Contributions}
In this paper, we propose a Generative Adversarial Network (GAN) and Large Language Model (LLM)-driven data augmentation framework for dynamic linguistic pattern modeling in Chinese sarcasm detection, and build a new Chinese sarcasm dataset \textit{SinaSarc}. Specifically, our contributions are summarized as follows.
\begin{enumerate}	
	\item \textbf{We present a novel GAN and LLM-driven data augmentation framework for Chinese sarcasm detection to dynamically model users’ linguistic patterns, which is the first time in this field.} 
	To address the scarcity of annotated Chinese sarcasm data, and discover hidden sarcastic cues from users' long-term linguistic patterns, we propose this framework. Specifically, We use the collected raw data to train the GANs to generate an annotated dataset enriched with user historical behavior, which allows the model to learn users’ long-term linguistic patterns and expressive styles,. Therefore, it can  better infer individual sarcasm tendencies and disambiguate otherwise ambiguous expressions. In addition, a GPT-3.5 based data augmentation technique is employed to enhance linguistic diversity and coverage. Compared to end-to-end LLM prompting, this hybrid design provides more structured and controllable data generation, while reducing the need for extensive prompting and repeated inference, resulting in a scalable and cost-efficient solution for constructing large-scale datasets. The experiment shows that our method outperforms the existing SOTA models and achieves the highest F1-scores on both the non-sarcastic and sarcastic categories, with values of 0.9138 and 0.9151 respectively.
	\item \textbf{SinaSarc: a new Chinese sarcasm dataset for dynamic users' linguistic pattern modeling.}
	Existing Chinese sarcasm datasets primarily focus on the comment and context, overlooking user-specific factors that shape sarcasm expression. To address this, we construct a new dataset, \textit{SinaSarc}, based on the proposed framework. This dataset contains 20,000 annotated instances and includes multi-dimensional attributes such as comment content, topic, comment hierarchy, and user historical behavior. By incorporating these behavior information, the dataset provides a foundation for accurately modeling users’ dynamic linguistic patterns.
	\item \textbf{We conduct extensive experiments to fully validate the effectiveness of our method.}
	Through comprehensive experimental evaluation, we verify that incorporating user historical behavior enables the model to capture users’ long-term linguistic patterns, allowing it to uncover deeper semantic cues in target comments. As a result, the model can disambiguate subtle expressions and reliably determine the sarcastic nature of comments, highlighting the critical role of user-level information in enhancing sarcasm detection.	
\end{enumerate}

\subsection{Organization}
This paper is organized into five sections. Section I introduces the research background and significance, summarizes existing studies on sarcasm detection, and outlines the main contributions of this paper. Section II reviews current methods for sarcasm detection, discussing their effectiveness and identifying existing challenges. In Section III, we present a new framework for Chinese sarcasm detection, with a detailed description of the overall design, dataset construction, and model development. Section IV describes the experimental setup and analyzes the results, demonstrating the effectiveness of the proposed method from multiple perspectives. Section V concludes the paper, providing a summary of the research as well as a discussion of its innovations and achievements.

\section{Related Work} 
In recent years, with the advancement of sentiment analysis techniques and the growing popularity of social media, research on sarcasm detection has received increasing attention. Existing sarcasm detection approaches can be categorized into three types: rule-based approaches, machine learning approaches, and deep learning approaches.

\subsection{Rule-Based Approaches}
Early sarcasm detection mainly relied on rule-based approaches that use predefined linguistic or emotional rules ~\cite{ref4, ref17, ref18, ref86, ref3, ref19}. Surve et al. ~\cite{ref4} observed that sarcasm on Twitter often appears as a contrast between positive sentiment words and negative situations. They proposed a bootstrapping algorithm that learns positive sentiment phrases and negative situation phrases from sarcastic tweets. Farias et al. ~\cite{ref17} treated sarcasm detection as a classification task and showed that emotional polarity and affective states are useful features. Veale et al. ~\cite{ref18} studied sarcasm in creative similes by detecting conceptual incompatibility between the attribute and the comparison object. There are also some methods used emoji polarity ~\cite{ref86}, emoticons and punctuation ~\cite{ref3}, and hashtags with polarity reversal rules ~\cite{ref19}. However, because sarcasm is often implicit and context-dependent, rule-based approaches have limited ability to capture subtle linguistic cues.

\subsection{Machine Learning Approaches}
Machine learning methods attempt to automatically learn features from data for sarcasm detection ~\cite{ref23, ref87, ref100, ref21}. Davidov et al. ~\cite{ref21} proposed a semi-supervised approach that extracts syntactic patterns and punctuation features from a small labeled dataset. Rodríguez et al. ~\cite{ref87} proposed SINCERE, a graph-based framework that uses sentiment and emotion embeddings, while Touahri et al. ~\cite{ref100} combined traditional machine learning models with neural models and sarcasm-specific lexical features. Although these methods improve performance compared with rule-based approaches, they rely heavily on manual feature engineering.

\subsection{Deep Learning Approaches}
Deep learning methods reduce the need for manual feature design and can learn complex semantic patterns. Existing approaches can be divided into target text-based approaches ~\cite{ref89, ref32, ref108, ref109, ref110, ref111} and context-based approaches ~\cite{ref29, ref114, ref115, ref116, ref117, ref118}.

Target text-based approaches mainly detect emotional inconsistency within the text. Poria et al. ~\cite{ref32} proposed a deep learning framework that uses pre-trained CNN models to extract sentiment, emotion, and personality features from tweets and combines them with semantic representations for classification. Ghosh et al. ~\cite{ref108} developed a hybrid neural model that integrates CNN, LSTM, and DNN to capture sequential dependencies and semantic information in short texts. Goel et al. ~\cite{ref110} proposed an ensemble framework that combines CNN, BiLSTM, and GRU with pre-trained embeddings such as Word2Vec, GloVe, and fastText. Recently, researchers have tended to focus on modeling internal relationships within the text. Tay et al. ~\cite{ref109} proposed the Multi-dimensional Intra-Attention Recurrent Network, which uses intra-attention to capture word-level contrast relationships while using LSTM to model sentence semantics. Wang et al. ~\cite{ref89} proposed KSDGCN, which constructs commonsense-augmented graphs and dependency graphs to capture emotional incongruity and entity relationships within the text.

Since sarcasm often depends on surrounding discourse, many studies incorporate contextual information. Zhang et al. ~\cite{ref29} proposed a dual-module BiLSTM model that captures both global and fine-grained sarcasm cues from context. Babanejad et al. ~\cite{ref116} introduced two models that combine contextual embeddings from BERT or SBERT with affective features to improve context understanding. Srivastava et al. ~\cite{ref117} proposed a hierarchical BERT architecture that encodes contextual utterances and target responses separately and models conversation-level dependencies. Potamias et al. ~\cite{ref118} further combined RoBERTa with a recurrent convolutional neural network to capture long-range contextual contradictions.

With the growth of social media, sarcasm detection has expanded from text-only analysis to multimodal analysis that incorporates images and other modalities ~\cite{ref85, ref90, ref95, ref96, ref97, ref98, ref88, ref91, ref93, ref94}. Wei et al. ~\cite{ref85} proposed DeepMSD, which extracts deep knowledge from text–image pairs using large vision–language models and performs cross-knowledge graph reasoning to detect incongruity. Wang et al. ~\cite{ref98} introduced the \(S^3\) Agent framework, which uses multiple analysis agents and large vision–language models to capture sarcastic cues from different perspectives. Liang et al. ~\cite{ref94} proposed MMGCL, which constructs a multimodal graph from textual, visual, and OCR features and applies graph contrastive learning to improve representation quality.

In Chinese sarcasm detection, current approaches have considered target text and contextual information but overlooked the relationship between users’ long-term linguistic styles and the emotional tendencies of their comments. In this paper, we present a novel GAN and LLM-driven data augmentation framework for Chinese sarcasm detection, which effectively learns users’ long-term linguistic patterns from their historical behavior to better identify the implicit sarcastic meanings behind commsents.	

\section{Methodology}
\begin{figure*}[htbp] 
	\includegraphics[width=\textwidth]{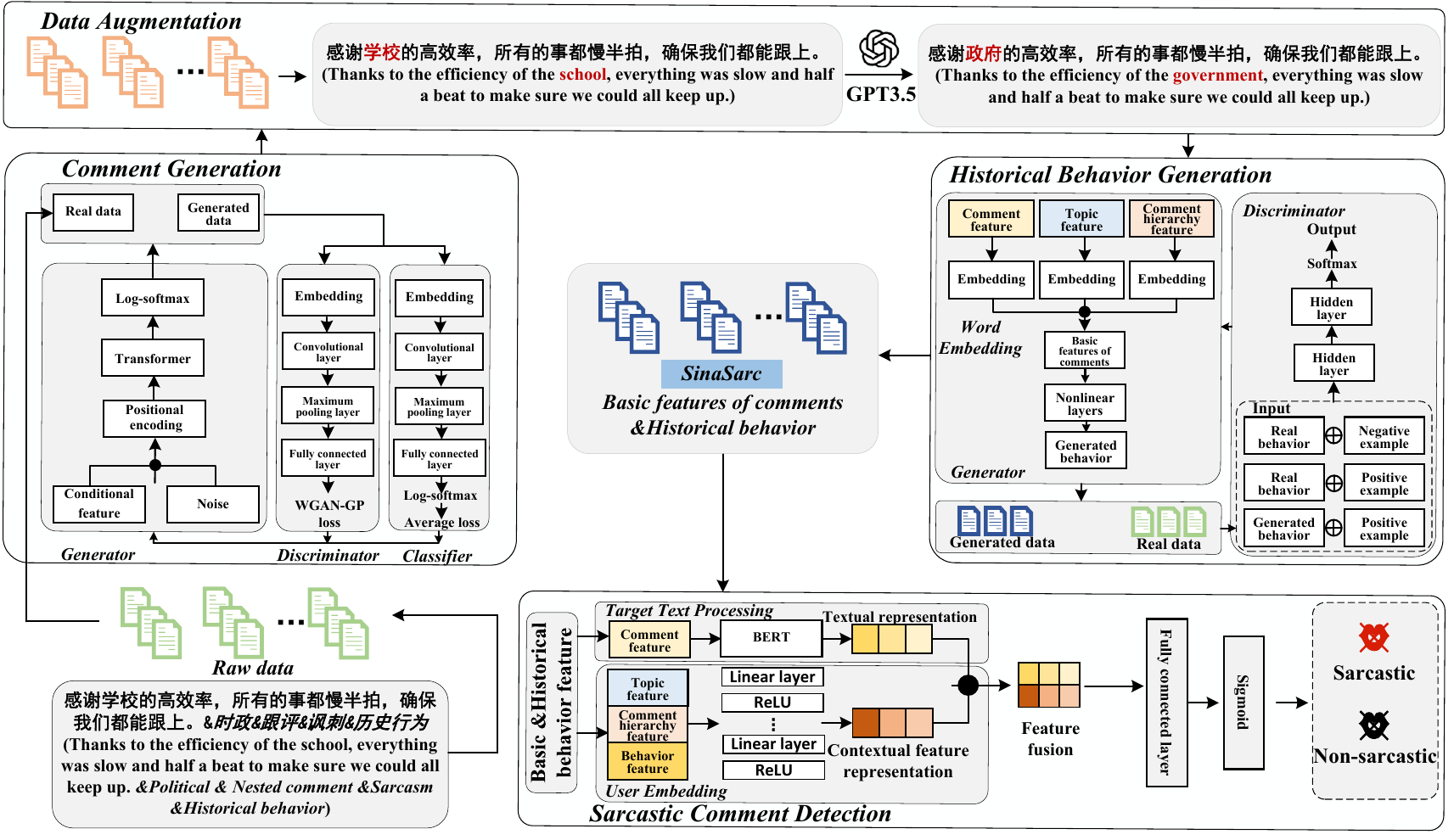}
	\caption{Modeling diagram of our proposed method.} 
	\label{fig1} 
\end{figure*}

In order to dynamically model users' long-term linguistic patterns, we proposes a framework that integrating user historical behavior for Chinese sarcasm detection, based on the generative adversarial networks and LLM-driven data augmentation technique, as shown in Fig.~\ref{fig1}. Specifically, our framework consists of 5 main modules: \textit{Dataset Construction, Comment Generation, Data Augmentation, Historical Behavior Generation, and Sarcastic Comment Detection}.

\subsection{Dataset Construction}
Most existing Chinese sarcasm datasets only include target text and contextual information. Few studies incorporate user historical behavior during dataset construction. Without such information, datasets cannot reflect users’ long-term linguistic patterns, which limits the modeling of individual emotional expression habits.

To address this issue, we first collected raw data, which was used to train the generation modules rather than the final detection model. We developed a web crawler to collect comments from Weibo across several topics, including \textit{lifestyle, politics, entertainment, relationships}, and \textit{public incidents}. These topics typically involve trending events and contain diverse user opinions. In addition to target comments, we also collected historical comments posted by those users.

After data collection, we removed noise such as advertisement links and HTML tags and standardized the text format. We collected some basic information, including \textit{Comment Content, Label, Topic, Comment Hierarchy}. To incorporate users’ linguistic patterns, we further extracted several historical behavior features for each user, including \textit{Comment Count, Topic Distribution, Sarcasm Rate, Comment Frequency, Reply Ratio}.

During the annotation stage, we first analyzed the linguistic characteristics of sarcastic comments and formulated corresponding annotation guidelines. In general, sarcasm in social media often exhibits properties such as contrast between literal and intended meanings, emotional reversal, targeted criticism, and strong dependence on contextual or cultural information. Data annotation was conducted by three annotators according to the predefined guidelines, resulting in a total of 5,000 annotated samples. Annotators were required to consider the overall context of the comment, including the related post and discussion background, and to make judgments based on the complete semantic meaning of the comment rather than isolated words or phrases. In addition, the annotation focused only on whether sarcasm was present, without considering the sentiment polarity of the comment. Given the subjective nature of sarcasm detection, we established three levels of sarcasm classification: 0 (sarcastic), 1 (non-sarcastic), and 2 (ambiguous). For comments labeled as 2 (ambiguous), group discussions were held to determine whether they contained sarcasm. The complete annotation guidelines and detailed examples are publicly available in our project repository on GitHub\footnote{https://github.com/yiyepianzhounc/SinaSarc}.

In the following work, we first train a GAN on the collected raw data to synthesize user comment information. We then apply GPT-3.5 for data augmentation to expand and diversify the dataset. Furthermore, considering that long-term linguistic patterns may encode implicit sarcasm cues, we employ an additional GAN to generate user historical behavior, enriching the dataset with user-level signals. This enables the detection model to not only learn textual and contextual semantics, but also capture users’ linguistic patterns for more accurate sarcasm identification.

By augmenting the raw data with the proposed framework, we constructed a dataset of 20,000 Chinese comments, named \textit{SinaSarc}. The dataset contains 10,000 sarcastic and 10,000 non-sarcastic instances, covering both target comments and user historical behaviors. 

Table~\ref{tab0} compares our dataset with existing Chinese sarcasm detection datasets. Most prior datasets, such as those from Tang et al.~\cite{ref103}, Lin et al.~\cite{ref104}, and Xiang et al.~\cite{ref119}, are constructed through manual annotation or lexicon-based approaches and mainly focus on textual features, with relatively limited consideration of contextual or user-level information. Some datasets (e.g., Gong et al.~\cite{ref107}, Zhang et al.~\cite{ref29}) incorporate contextual information; however, the distribution of sarcastic and non-sarcastic samples remains relatively imbalanced, which may affect the learning of robust sarcasm patterns. In contrast, our \textit{SinaSarc} dataset is built on a GAN and LLM-driven data augmentation framework, which introduces diverse and semantically consistent samples. It contains 20,000 instances evenly distributed between sarcastic and non-sarcastic classes, providing a balanced setting for model training. In addition, the dataset incorporates user historical behavior as an integral component, enabling models to capture users’ long-term linguistic patterns, which are important for understanding the implicit and context-dependent nature of sarcasm.

\begin{table*}[htbp]
	\centering
	\caption{Chinese Sarcasm Dataset Comparison}
	\label{tab0}
	\begin{tabular}{
			l 
			>{\centering\arraybackslash}m{1.4cm} 
			>{\centering\arraybackslash}m{1.1cm} 
			>{\centering\arraybackslash}m{1.4cm} 
			>{\centering\arraybackslash}m{1.8cm} 
			>{\centering\arraybackslash}m{3.5cm}
			>{\centering\arraybackslash}m{4cm} 
		}
		\toprule
		Dataset & Source & Total & Sarcastic & Non-sarcastic & Method & Key Components \\
		\midrule
		Tang et al.~\cite{ref103} & Plurk & 950 & 950 & 0 & Bootstrapping+Manual & Text \\
		Lin et al.~\cite{ref104} & PTT & 26,629 & 17,256 & 9,373 & Lexicon-based & Text \\
		Gong et al.~\cite{ref107} & Guanchazhe & 91,782 & 2,486 & 89,296 & Manual & Contextual Information \\
		Xiang et al.~\cite{ref119} & Sina Weibo & 8,702 & 968 & 7,734 & Manual & Text \\
		Zhang et al.~\cite{ref29} & Bilibili & 79,045 & 2,814 & 76,231 & Manual & Contextual Information \\
		SinaSarc (Ours) & Sina Weibo & 20,000 & 10,000 & 10,000 & Data Augmentation & User Historical Behavior \\
		\bottomrule
	\end{tabular}
\end{table*}

\subsection{Comment Generation}
As shown in Fig.~\ref{fig1}, this module generates several Chinese comments with category labels (sarcastic, non-sarcastic) and contextual information. To ensure the generated comments are semantically coherent, highly realistic and consistent with target attributes, we construct a generative framework based on Transformer ~\cite{ref120} decoder and Wasserstein Generative Adversarial Network with Gradient Penalty (WGAN-GP) ~\cite{ref121}, which integrates a pre-trained BERT ~\cite{ref81} embedding layer, a Transformer generator, a Text-CNN ~\cite{ref122} based WGAN-GP discriminator and a sentiment classifier. The whole framework is trained in two stages: generator pre-training and joint adversarial training.

The generator takes random noise and conditional feature vectors as inputs to generate comment sequences with specified attributes. The discriminator distinguishes real comments from synthetic ones to enhance the authenticity of outputs. The classifier constrains the sentiment consistency of generated comments.

\begin{itemize}
	\item \textbf{Generator}
\end{itemize}

The generator adopts a Transformer decoder as the core architecture, which is used to model the sequential dependency of comments and realize autoregressive generation. It takes random latent noise $z\in\mathbb{R}^{128} $ and conditional feature vector $f\in\mathbb{R}^{9}$ as inputs, where the conditional feature is formed by concatenating sentiment label, topic and comment hierarchy.

First, the noise and conditional feature are fused and projected to generate the memory vector for decoding, as shown in Equation~\ref{eq:gen_memory}.
\begin{equation}
	\text{memory} = \sigma\left(W_{proj}\cdot\left[z;f\right]+b_{proj}\right)
	\label{eq:gen_memory}
\end{equation}
where $[z;f]$ denotes concatenation, $W_{proj}$ and $b_{proj}$ are trainable parameters, and $\sigma$ denotes the ReLU activation function.

The generation process starts with the special token $\langle \text{SOS} \rangle$, and generates each token autoregressively with the guidance of positional encoding and Transformer decoder layers. As shown in Equation~\ref{eq:gen_token}, the output layer maps decoder states to vocabulary distribution via log-softmax.
\begin{equation}
	\begin{split}
	P(w_t|w_1,\dots,w_{t-1},z,f,\theta_G) = 
	\\
	\text{LogSoftmax}\left(W_{out}\cdot H_t + b_{out}\right)
	\end{split}
	\label{eq:gen_token}
\end{equation}
where $H_t$ is the hidden state at step $t$ output by the Transformer decoder, and $\theta_G$ denotes all parameters of the generator.

During pre-training, the generator is optimized by minimizing the negative log-likelihood loss between generated sequences and real sequences, as shown in Equation~\ref{eq:gen_pre_loss}.
\begin{equation}
	\begin{split}
		\mathcal{L}_{G_{pre}} =& -\frac{1}{N}\sum_{i=1}^{N}\sum_{t=1}^{T}\mathbb{I}(w_{i,t}\neq \text{pad}) \cdot \\
		& \log P(w_{i,t}|w_{i,1},\dots,w_{i,t-1},z,f,\theta_G)
	\end{split}
	\label{eq:gen_pre_loss}
\end{equation}
where $N$ is batch size, $T$ is maximum sequence length, and $\mathbb{I}(\cdot)$ is the indicator function to mask padding tokens.

In adversarial training, the generator is optimized by a weighted combination of adversarial loss and classification loss, as shown in Equation~\ref{eq:gen_adv_loss}:
\begin{equation}
	\mathcal{L}_G = \alpha \cdot \mathcal{L}_{G_{adv}} + (1-\alpha) \cdot \mathcal{L}_{G_{cls}}
	\label{eq:gen_adv_loss}
\end{equation}
where $\alpha \in (0,1)$ is the weight hyperparameter to balance the authenticity and attribute consistency of generated comments.

\begin{itemize}
	\item \textbf{Discriminator}
\end{itemize}

The discriminator adopts a multi-filter Text-CNN structure to judge the authenticity of input comment sequences. It takes comment sequence $S$ and conditional feature $f$ as inputs, and outputs a realism score without activation.

First, the word embedding of sequence and conditional feature are concatenated along the feature dimension, as shown in Equation~\ref{eq:dis_emb_concat}:
\begin{equation}
	X = \left[\text{Emb}(S);f\otimes\mathbf{1}_T\right]
	\label{eq:dis_emb_concat}
\end{equation}
where $\text{Emb}(S)$ is the word embedding matrix of sequence $S$, and $\mathbf{1}_T$ denotes a all-one vector for feature expansion.

Multi-scale convolution and max-pooling are applied to extract local text features, which is described in Equation~\ref{eq:dis_conv}:
\begin{equation}
	F_k = \text{MaxPool}\left(\delta\left(\text{Conv2D}_k(X)\right)\right)
	\label{eq:dis_conv}
\end{equation}
where $\text{Conv2D}_k$ denotes convolution with kernel size $k$, $\delta$ denotes ReLU activation, and $\text{MaxPool}$ denotes adaptive max-pooling.

The concatenated features are fed into a fully connected layer to obtain the final score, as presented in Equation~\ref{eq:dis_score}:
\begin{equation}
	D(S,f,\theta_D) = W_{fc} \cdot \left[F_k\right] + b_{fc}
	\label{eq:dis_score}
\end{equation}
where $\left[F_k\right]$ denotes the concatenation of multi-scale local feature vectors, and $\theta_D$ denotes all trainable parameters of the discriminator.

The WGAN-GP loss of discriminator is defined as Equation~\ref{eq:dis_loss}:
\begin{equation}
	\mathcal{L}_D = \mathbb{E}_{S\sim P_{real}}[D(S,f)] - \mathbb{E}_{S\sim P_{fake}}[D(S,f)] + \lambda_{GP}\cdot\mathcal{L}_{GP}
	\label{eq:dis_loss}
\end{equation}
where $\mathbb{E}[\cdot]$ denotes the expectation operator, $S\sim P_{real}$ indicates that the sample $S$ is drawn from the real user comment distribution, $S\sim P_{fake}$ indicates that the sample $S$ is drawn from the synthetic comment distribution generated by the generator. $\lambda_{GP}$ is the gradient penalty coefficient, and $\mathcal{L}_{GP}$ is the gradient penalty term that enforces the Lipschitz constraint on the discriminator.

\begin{itemize}
	\item \textbf{Classifier}
\end{itemize}

The classifier shares the same Text-CNN feature extraction structure with the discriminator, and is used to predict the sentiment label of input comments, so as to guide the generator to produce label-consistent outputs. It takes sequence $S$ and conditional feature $f$ as inputs and outputs the log-probability of sentiment labels via LogSoftmax, as shown in Equation~\ref{eq:cls_output}:
\begin{equation}
	C(S,f,\theta_C) = \text{LogSoftmax}\left(W_{cls}\cdot H_{feat} + b_{cls}\right)
	\label{eq:cls_output}
\end{equation}
where $\theta_C$ denotes parameters of the classifier, and $H_{feat}$ is the fused Text-CNN feature.

The classification loss is the average of negative log-likelihood loss on real comments and synthetic comments, as formulated in Equation~\ref{eq:cls_loss}:
\begin{equation}
	\mathcal{L}_C = \frac{1}{2}\left(\mathcal{L}_{C_{real}} + \mathcal{L}_{C_{fake}}\right)
	\label{eq:cls_loss}
\end{equation}
where $\mathcal{L}_{C_{real}}$ and $\mathcal{L}_{C_{fake}}$ are computed by Equation~\ref{eq:cls_real_loss} and Equation~\ref{eq:cls_fake_loss} respectively:
\begin{equation}
	\mathcal{L}_{C_{real}} = -\mathbb{E}_{(S,f,y)\sim P_{real}}\left[C(S,f,\theta_C)[y]\right]
	\label{eq:cls_real_loss}
\end{equation}
\begin{equation}
	\mathcal{L}_{C_{fake}} = -\mathbb{E}_{(S,f,y)\sim P_{fake}}\left[C(S,f,\theta_C)[y]\right]
	\label{eq:cls_fake_loss}
\end{equation}
where $y$ denotes the ground-truth sentiment label.

This module is capable of generating comments with contextual information. These comments follow real language patterns and possess specific topic and comment hierarchy, thereby providing high-quality synthetic data for subsequent tasks.

\subsection{Data Augmentation}
Constructing a large, manually labeled dataset is costly, which makes data augmentation a preferred approach. Data augmentation typically relies on invariance, rules, or heuristic knowledge. This study addresses the challenges of data scarcity and limited diversity in Chinese sarcasm comment detection through data augmentation techniques. By applying data augmentation, this research generates new and diverse samples, improving the model’s generalization ability and robustness to various expressions of sarcasm. Data augmentation is mainly performed through context-aware text replacement with GPT-3.5. 

The text replacement strategy generates new comments by identifying and replacing key words within the sentences, as shown in Equation~\ref{eq8}.
\begin{equation} 
	T\mathrm{'}=T-W+W\mathrm{'}
	\label{eq8} 
\end{equation}
In this context, $T$ represents the original text, $W$ denotes the selected vocabulary for replacement, $W\mathrm{'}$ refers to the new vocabulary after replacement, and $T\mathrm{'}$ is the generated new text. To ensure coherence and maintain semantic integrity, the replaced vocabulary is typically selected as synonyms or contextually similar words. Previous studies on data augmentation through text replacement primarily used synonyms for target words. However, synonyms are often limited and lack sufficient diversity, which can result in inadequate data variation after replacement. Therefore, this study employs GPT-3.5 to find contextually similar words for replacement, which maintains the natural fluency of sentences.

\begin{CJK}{UTF8}{gbsn}
	We applies ChatGPT to perform context-aware word replacement.  We leverage deep contextual representations to predict semantically appropriate alternatives for target words, expanding beyond traditional synonym substitution. For example, in the sarcastic statement, ``这张票真值了，两个小时的电影感觉看了五个小时" (This movie ticket was totally worth it—two hours that felt like five!), a synonym replacement approach might substitute ``电影" (movie) with ``影片" (film). In contrast, the LLM can generate more diverse yet contextually coherent replacements such as ``戏剧" (drama), ``音乐会" (concert) or ``脱口秀" (talk show), while preserving the sarcastic tone. 
\end{CJK}

\subsection{Historical Behavior Generation}
For our generated data, we focus on comment texts, topic information, and comment hierarchy, similar to existing research on Chinese sarcasm detection. However, the impact of mining user linguistic patterns on sarcasm detection remains under-explored in this field.

As shown in Fig.~\ref{fig1}, we developed a model based on GAN to generate user historical behavior features ~\cite{ref123}, which reflects users’ linguistic patterns. This model aims to learn the mapping from users' basic comment features to their historical behaviors based on the raw data we collected, and then synthesize realistic historical behavior features for the generated data in our work above.

This model consists of two key components: the generator and the discriminator. The generator takes the basic comment features of input comments and generates corresponding historical behavior features. We selected three types of basic comment features: comment content, topics, and comment hierarchy (top-level comment, nested comment). The discriminator distinguishes between the generated historical behavior features and the real behavior features, thereby improving the training of the generator.

\textit{\textbf{a) Feature Extraction:}}
We extract two types of features from the dataset: basic comment features and user historical behavior features.

\begin{itemize}
	\item \textbf{Basic Comment Features}
\end{itemize}

\begin{enumerate}
	\item \textbf{Comment Content:} The textual content of the comment, which captures the basic semantic information of the comment. \cite{ref11,ref12}.
	
	\item \textbf{Topic:} The topic associated with the comment. Since Weibo comments are posted under specific posts or themes, topic information may influence the occurrence of sarcasm \cite{ref15}.
	
	\item \textbf{Comment Hierarchy:} Whether the comment is a top-level comment or a nested one. Nested comments often appear in discussions and interactions, where sarcastic expressions are more likely to occur.
\end{enumerate}

\begin{itemize}
	\item \textbf{User Historical Behavior Features}
\end{itemize}

We extract five dimensions of user historical behavior features:

\begin{enumerate}
	\item \textbf{Comment Count:} The total number of comments previously posted by the user.
	
	\item \textbf{Topic Distribution:} The main topics that the user has commented on in the past, which may relate to sarcasm tendencies \cite{ref23, ref8}.
	
	\item \textbf{Sarcasm Rate:} The proportion of sarcastic comments in the user's historical comments, reflecting the user's habitual use of sarcasm \cite{ref23,ref8}.
	
	\item \textbf{Comment Frequency:} The average daily comment count of the user over a given time span.
	
	\item \textbf{Reply Ratio:} The ratio of nested comments to the total number of historical comments, indicating the user's level of interaction in discussions \cite{ref23}.
\end{enumerate}

\textbf{\textit{b) Generator:}}
Before introducing the model, we provides a standard explanation of the abbreviations used for the reader’s convenience, PE for Positive Example, NE for Negative Example, GB for Generated Behavior, and RB for Real Behavior. 

The first three layers of the generator capture the comment content, topic, and comment hierarchy. A nonlinear hidden layer then transforms the basic comment features into generated historical behavior features. These generated features should align with the basic features of the input comments and closely resemble real historical behavior. The relevant mathematical expressions are as follows:
\begin{equation} 
		L_t=\underset{\varTheta G}{min}J(D(PE\oplus GB),1)
\end{equation}
\begin{equation} 
		L_c=\underset{\varTheta G}{min}J(RB,GB)
\end{equation}
\begin{equation} 
		L_G=\lambda L_t + (1-\lambda )L_c
\end{equation}
Here, $L_t$ represents the loss in the process of generating historical behavior features from basic comment features. $L_c$ denotes the closeness loss, which ensures that the generated historical behavior features are more realistic and reasonable. $L_G$ is the total loss of the generator. $J$ represents the cross-entropy function. $D$ is the discriminator function. The $\oplus$ symbol indicates the connection of two embeddings. The balancing parameter $\lambda$ in $L_G$ controls the weights of the two losses.

\textbf{\textit{c) Discriminator:}}
As shown in Fig.~\ref{fig1}, the discriminator is used to determine whether the generated historical behavior features are similar to the real historical behavior features. The goal of the discriminator is to guide training by distinguishing between generated and real behavior features. Specifically, the discriminator classifies pairs from real training data (PE, RB) as true, pairs from the generator (PE, GB) as false, and unmatched fake samples (NE, RB) as false. The relevant mathematical expressions are as follows:
\begin{equation}
		L_r=\underset{\varTheta D}{min}J(D(PE\oplus RB),1)
\end{equation}
\begin{equation} 
		L_f=\underset{\varTheta D}{min}J(D(PE\oplus GB),0)
\end{equation}
\begin{equation}
		L_h=\underset{\varTheta D}{min}J(D(NE\oplus RB),0)
\end{equation}
\begin{equation}
		L_D = L_r + \frac{1}{2}L_f + \frac{1}{2}L_h
\end{equation}
This paper defines four loss functions: $L_r$ to measure the discernibility of real samples, $L_f$ to measure the discernibility of generated samples, and $L_h$ to measure the discernibility of fake samples. $L_D$ represents the total loss of the discriminator.

\subsection{Sarcastic Comment Detection}
As shown in Fig.~\ref{fig1}, this paper proposes a Chinese sarcasm detection model that incorporates user historical behavior to learn their linguistic patterns. The model consists of three sub-modules: the target text processing, the user embedding, and the feature fusion. This model combines text information with user historical behavior to dynamically model users’ linguistic patterns.

In the target text processing module, the BERT is used to process the comment content and capture semantic information. The user embedding module uses a fully connected neural network to process context and user historical behavior, extracting relevant users' linguistic patterns. The feature fusion module combines the outputs from the target text processing and user embedding modules, connecting the two sets of features while sharing layer parameters to capture the combined representation of both text and users' linguistic patterns.

\subsubsection{Target Text Processing}
The text processing module is implemented by fine-tuning a pre-trained BERT model. This paper modifies the BERT configuration based on task requirements, adjusting the number of hidden layers and attention heads to fit the sarcasm detection task. The module represents the input comment content as input IDs and attention masks. Through the BERT encoding process, the last hidden state corresponding to the [CLS] token is extracted as the representation of the comment, denoted as $H\in \mathbb{R} ^{L\times d}$. This captures the semantic information within the comment to help determine if it contains sarcasm, as shown in Equation~\ref{eq16}.
\begin{equation} 
		H=BERT(E,M)
		\label{eq16}
\end{equation}
The $BERT$ function represents the encoding process of the BERT model. $E$ denotes the embedding representation of the input sequence $X$, specifically $E=[e_1,e_2,...,e_n]$, where $e_i$ is the embedding representation of the $i$-th token in the input sequence $X=[x_1,x_2,...,x_n]$, and $x_i$ is the $i$-th token in the input sequence. $M$ indicates the attention mask for the input sequence, representing valid positions.

\subsubsection{User Embedding}
The user embedding module processes context and user historical behavior. These information are represented as a numeric feature vector, including \textit{Label, Topic, Comment Hierarchy, Comment Count, Topic Distribution, Sarcasm Rate, Comment Frequency, Reply Ratio}. This module uses a fully connected neural network composed of multiple linear layers and activation functions to map the user embedding vector $x\in \mathbb{R} ^k$ to a fixed-dimensional user feature representation $u\in \mathbb{R} ^m$. The embedded user features reflect users' linguistic patterns, providing auxiliary information for sarcasm detection. The mathematical expression is shown in Equation~\ref{eq17}.
\begin{equation}  
		u=f(x)=max\!\:(0,W^Tx+b)
		\label{eq17} 
\end{equation}
Where $f(x)$ represents the ReLU activation function, $W$ denotes the weights of the multiple linear layers, and $b$ is the bias term.

\subsubsection{Feature Fusion}
The feature fusion module concatenates the outputs of the target text processing and user embedding modules. It uses a fully connected neural network with shared layer parameters to classify the comments. This study employs a fully connected neural network as the classifier, consisting of linear layers and a Sigmoid activation function. The concatenated features are input to the network, which outputs the probability that the comment is sarcastic. The text information captures the semantics within the comments, while the user historical behavior reflects the users' linguistic patterns. The combination of these two information sources aids in better understanding the comment's meaning and the users' intent. The mathematical expressions are as follows:
\begin{equation} 
		combined=[H,u]
\end{equation}
\begin{equation}
		z=W_{cls}\cdot combined+b_{cls}
\end{equation}
\begin{equation} 
		y=sigmoid\left( z \right)
\end{equation}
Where $H\in \mathbb{R} ^{L\times d}$ represents the output of the text processing module, $u\in \mathbb{R} ^m$ is the output of the user embedding module, $W_{cls}$ denotes the weights of the fully connected neural network with shared layer parameters, $b_{cls}$ is the bias term, and $\cdot$ indicates matrix multiplication. The Sigmoid function maps the output to a probability value between 0 and 1.

\section{Experiments}

\subsection{Experiment settings}
\subsubsection{Experimental Environment}
In this section, we conducted various experiments to demonstrate the effectiveness of our proposed sarcasm detection method. All experiments were carried out in an environment with an Intel(R) Xeon(R) Gold 6130 CPU @ 2.10GHz processor and Tesla V100 GPU with 32GB memory. 

\subsubsection{Experimental Data}
This experiment used the proposed \textit{SinaSarc} dateset. The dataset is balanced, containing 10,000 sarcastic comments and 10,000 non-sarcastic comments, for a total of 20,000 samples. Each sample includes the comment content, the corresponding label, topic, comment hierarchy, and user historical behavior. The balanced dataset helps avoid bias toward any particular category during model training, ensuring better performance in sarcasm detection. The dataset was split into training, validation, and test sets with a 6:2:2 ratio. We used \textit{Accuracy (Acc.),  Precision (Pre.), Recall (Rec.)}, and \textit{F1 Score} to evaluate the performance of our detection method.

\subsection{Comparison Experiments}
To demonstrate that the model proposed in this study outperforms existing models in the Chinese sarcasm detection task, this section compares the proposed model with SOTA text classification models, including machine learning models \textit{Random Forest, SVM, AdaBoost}, neural network models \textit{FNN, LSTM, BiLSTM, LSTM-Att, BiLSTM-Att}, pre-trained transformer models \textit{BERT-base, BERT-large, RoBERTa-base, RoBERTa-large, SBERT}, and large language models (LLMs)\textit{ GPT-4-Turbo, Qwen2-7B, Baichuan2-7B-Chat, Gemini-1.5-Pro, Mixtral 8x7B}. We also assessed the impact of extracted text features (T), contextual features (C), and user historical behavior features (U) on sarcasm detection.

Table~\ref{tab4} shows the performance of various baseline models and our proposed model on the dataset. It is clear that models focusing solely on the target text perform poorly. Models that consider contextual information show slightly better performance, but none of these models account for the impact of users' linguistic patterns on sarcasm detection. Our model performs the best, achieving the highest F1 score and accuracy. This is because our model integrates text features, contextual information, and user historical behavior features. The multi-dimensional feature fusion effectively captures the complexity of sarcastic comments and user behavior habits, providing deeper insights into user linguistic patterns and preferences.

The second-best performance comes from the RoBERTa-large model. This indicates that the RoBERTa-large effectively captures information in the text, enhancing the understanding of sentence structure and better incorporating user historical behavior than other models. Models only focusing on target comment text features perform poorly. This may be because these traditional models may not fully capture the subtle variations and deeper semantics of sarcasm.

\subsection{Ablation Experiment}
To validate the effectiveness of the user historical behavior features proposed in this paper, we conducted ablation experiments on various features. The subsets of the feature set can be represented using the set difference function, as shown in Equation~\ref{eq21}.
\begin{equation} 
	F \setminus F' = \{x \mid x \in F \land x \notin F'\}
	\label{eq21} 
\end{equation}
Where $F$ represents the overall collection of user historical behavior features, and $F\mathrm{'}$ represents a subset of the user historical behavior features collection. In the ablation experiments, we selected the following targets for ablation:  

\begin{itemize}
	\item $F$: The complete set of user historical behavior features.
	\item $F \setminus CC$: The user historical behavior feature set without \textit{Comment Count}.
	\item $F \setminus TD$: The user historical behavior feature set without \textit{Topic Distribution}.
	\item $F \setminus SR$: The user historical behavior feature set without \textit{Sarcasm Rate}.
	\item $F \setminus CF$: The user historical behavior feature set without \textit{ Comment Frequency}.
	\item $F \setminus RR$: The user historical behavior feature set without \textit{Reply Ratio}.
\end{itemize}

\begin{table*}[htbp]
	\centering
	\caption{Results of comparison of experimental}
	\label{tab4}
	\begin{tabularx}{\textwidth}{
			>{\centering\arraybackslash}p{1.7cm}
			>{\centering\arraybackslash}p{4cm}
			*{3}{>{\centering\arraybackslash}p{0.4cm}}
			>{\centering\arraybackslash}p{1.2cm}
			*{6}{>{\centering\arraybackslash}p{0.85cm}}
		}
		\toprule
		\multirow{2}{*}{\textbf{Category}} & \multirow{2}{*}{\textbf{Model}} & \multicolumn{3}{c}{\textbf{Features}} & \multirow{2}{*}{\textbf{Acc.}} & \multicolumn{3}{c}{\textbf{Non-sarcastic}} & \multicolumn{3}{c}{\textbf{Sarcastic}} \\
		\cmidrule(lr){3-5} \cmidrule(lr){7-9} \cmidrule(lr){10-12}
		& & \textbf{T} & \textbf{C} & \textbf{U} & & \textbf{Pre.} & \textbf{Rec.} & \textbf{F1} & \textbf{Pre.} & \textbf{Rec.} & \textbf{F1} \\
		\midrule
		\multirow{3}{*}{Machine Learning} 
		& Random Forest ~\cite{ref76} & \checkmark & & & 0.8032 & 0.7985 & 0.8912 & 0.8423 & 0.8122 & 0.6766 & 0.7383 \\
		& SVM ~\cite{ref29} & \checkmark & & & 0.8183 & 0.8569 & 0.8307 & 0.8436 & 0.7668 & 0.8005 & 0.7833 \\
		& AdaBoost ~\cite{ref77} & \checkmark & & & 0.7231 & 0.7506 & 0.7946 & 0.7720 & 0.6775 & 0.6204 & 0.6477 \\
		\midrule
		\multirow{5}{*}{Neural Network} 
		& FNN ~\cite{ref78} & \checkmark & & & 0.8464 & 0.8610 & 0.8820 & 0.8714 & 0.8242 & 0.7953 & 0.8095 \\
		& LSTM ~\cite{ref109} & \checkmark & \checkmark & & 0.8428 & 0.8650 & 0.8692 & 0.8671 & 0.8106 & 0.8049 & 0.8078 \\
		& BiLSTM ~\cite{ref84} & \checkmark & \checkmark & & 0.8371 & 0.8545 & 0.8722 & 0.8633 & 0.8107 & 0.7865 & 0.7984 \\
		& LSTM-Att ~\cite{ref79} & \checkmark & \checkmark & & 0.8335 & 0.8515 & 0.8692 & 0.8603 & 0.8062 & 0.7821 & 0.7939 \\
		& BiLSTM-Att ~\cite{ref80} & \checkmark & \checkmark & & 0.8414 & 0.8598 & 0.8735 & 0.8666 & 0.8138 & 0.7953 & 0.8044 \\
		\midrule
		\multirow{5}{*}{Pre-trained} 
		& BERT-base ~\cite{ref81} & \checkmark & \checkmark & \checkmark & 0.8627 & 0.8680 & 0.9046 & 0.8860 & 0.8541 & 0.8023 & 0.8274 \\
		& BERT-large ~\cite{ref81} & \checkmark & \checkmark & \checkmark & 0.8735 & 0.8804 & \underline{0.9089} & 0.8944 & 0.8627 & 0.8225 & 0.8421 \\
		& RoBERTa-base ~\cite{ref101} & \checkmark & \checkmark & \checkmark & 0.8800 & \underline{0.8922} & 0.9059 & 0.8990 & 0.8616 & 0.8427 & 0.8521 \\
		& RoBERTa-large ~\cite{ref101} & \checkmark & \checkmark & \checkmark & \underline{0.8861} & 0.8915 & \textbf{0.9187} & \underline{0.9049} & \underline{0.8778} & 0.8392 & \underline{0.8580} \\
		& SBERT ~\cite{ref102} & \checkmark & \checkmark & \checkmark & 0.8298 & 0.8411 & 0.8771 & 0.8588 & 0.8118 & 0.7619 & 0.7860 \\
		\midrule
		\multirow{5}{*}{LLMs} 
		& GPT-4-Turbo\footnotemark[2] & \checkmark & \checkmark & \checkmark & 0.8378 & 0.8638 & 0.8606 & 0.8622 & 0.8008 & 0.8050 & 0.8029 \\
		& Qwen2-7B\footnotemark[3] & \checkmark & \checkmark & \checkmark & 0.8574 & 0.8480 & 0.8710 & 0.8593 & 0.8674 & \underline{0.8438} & 0.8554 \\
		& Baichuan2-7B\footnotemark[4] & \checkmark & \checkmark & \checkmark & 0.8578 & 0.8798 & 0.8790 & 0.8794 & 0.8263 & 0.8274 & 0.8269 \\
		& Gemini-1.5-Pro\footnotemark[5] & \checkmark & \checkmark & \checkmark & 0.8337 & 0.8458 & 0.8780 & 0.8616 & 0.8146 & 0.7701 & 0.7917 \\
		& Mixtral 8x7B\footnotemark[6] & \checkmark & \checkmark & \checkmark & 0.8332 & 0.8433 & 0.8808 & 0.8616 & 0.8170 & 0.7648 & 0.7901 \\
		\midrule
		\textbf{Proposed} & \textbf{Our model} & \checkmark & \checkmark & \checkmark & \textbf{0.9144} & \textbf{0.9208} & 0.9069 & \textbf{0.9138} & \textbf{0.9083} & \textbf{0.9220} & \textbf{0.9151} \\
		\bottomrule
	\end{tabularx}
\end{table*}
\footnotetext[2]{OpenAI, 2024. GPT-4-Turbo introduction. https://openai.com/index/hello-gpt-4o/}
\footnotetext[3]{https://huggingface.co/Qwen/Qwen-7B}
\footnotetext[4]{https://huggingface.co/baichuan-inc/Baichuan2-7B-Base}
\footnotetext[5]{Google DeepMind, 2025. https://blog.google/technology/ai/google-gemini-next-generation-model-february-2025/}
\footnotetext[6]{Mistral AI, 2023. Mixtral of experts. https://mistral.ai/news/mixtral-of-experts}

\begin{figure}[ht]
	\centering
	\includegraphics[width=2.4in]{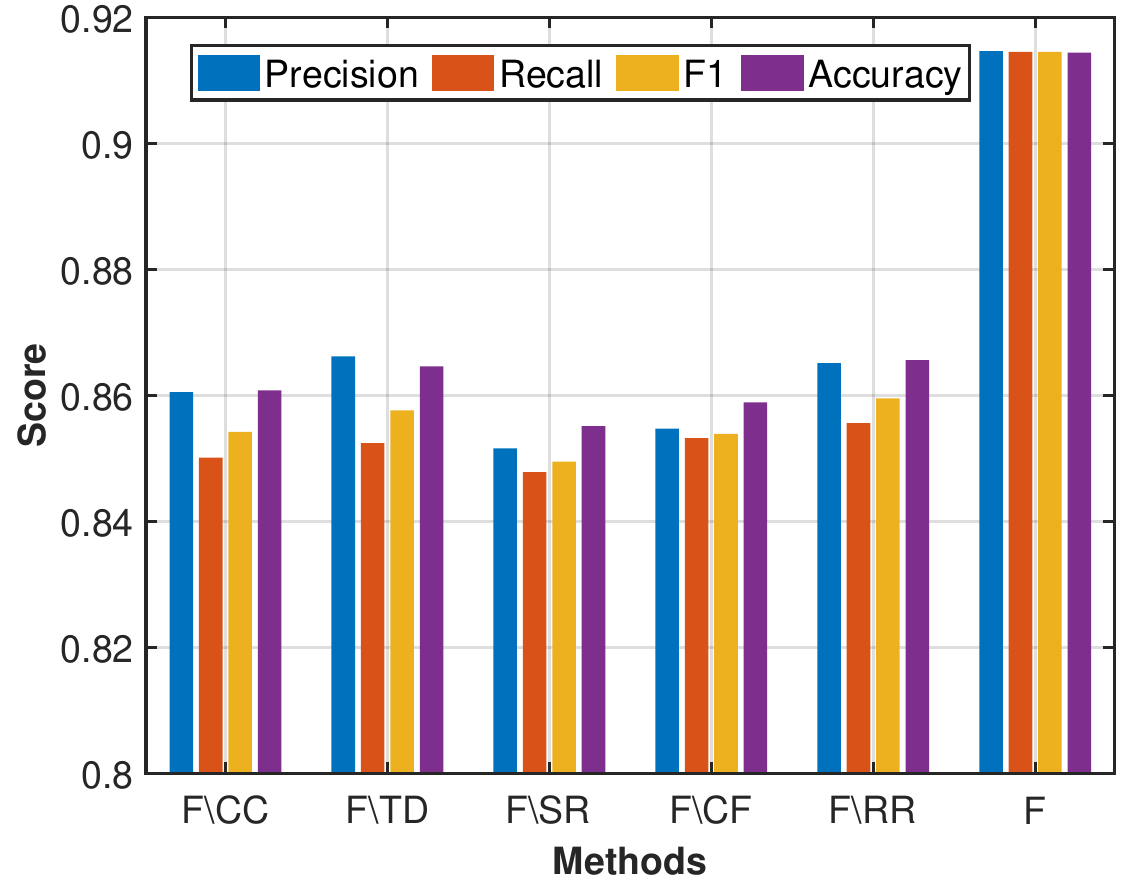}
	\caption{Results of ablation experiments}
	\label{fig2}
\end{figure}

The experimental results are shown in Fig.~\ref{fig2}. It is evident that detection performance decreases when a specific behavioral feature is removed, indicating that the selected user historical behavior features are effective for Chinese sarcasm detection. Upon further analysis, when the feature \textit{Sarcasm Rate} is excluded, We observed that the model's performance declined the most. This phenomenon may be attributed to the strong correlation between sarcasm and individual language habits: the same statement can convey different meanings depending on the speaker.

When the \textit{Sarcasm Rate} feature is absent, the model relies more heavily on textual cues. For example, the comment \begin{CJK}{UTF8}{gbsn}``这种真的是天才"
\end{CJK}(``This is truly genius") may appear neutral or even positive. However, when made by a user with a high historical tendency toward sarcasm, the full model correctly identifies it as sarcastic. In contrast, without this feature (as in the `F/SR' setting), the model behaves more conservatively and only labels comments as sarcastic when highly confident. Since the semantics of this comment are not overtly sarcastic, the model classifies it as non-sarcastic, leading to a decline in model performance.

\subsection{Noise Experiment}

To assess the model's performance when handling noisy data, we designed a noise experiment. This experiment simulates common imperfections in real-world data processing by artificially introducing label noise. We randomly selected a certain percentage of data samples (from 0.05 to 0.45) in the training set and changed their labels to the opposite ones. Let $L$ be the original label set, and we define a noise function $f$ that alters the labels with a certain probability $p$, where $p$ ranges from 0.05 to 0.45. The noisy label set is $L\mathrm{'}$:
\begin{equation} 
		L\mathrm{'}=f(L,p)
\end{equation}
In this equation, the function $f$ operates on each label $l \in L$ and changes $l$ to its opposite with probability $p$. For instance, if the original label is ``sarcastic," it will be changed to ``non-sarcastic" with probability $p$, and vice versa.

In the experiment, we first trained the baseline models without noise to establish a performance benchmark. Then, the models were trained on training data with varying levels of noise, with independent training and evaluation conducted for each noise level to observe the specific effects of noise on model performance.

\begin{figure}[htbp]
	\centering
	\includegraphics[width=2.4in]{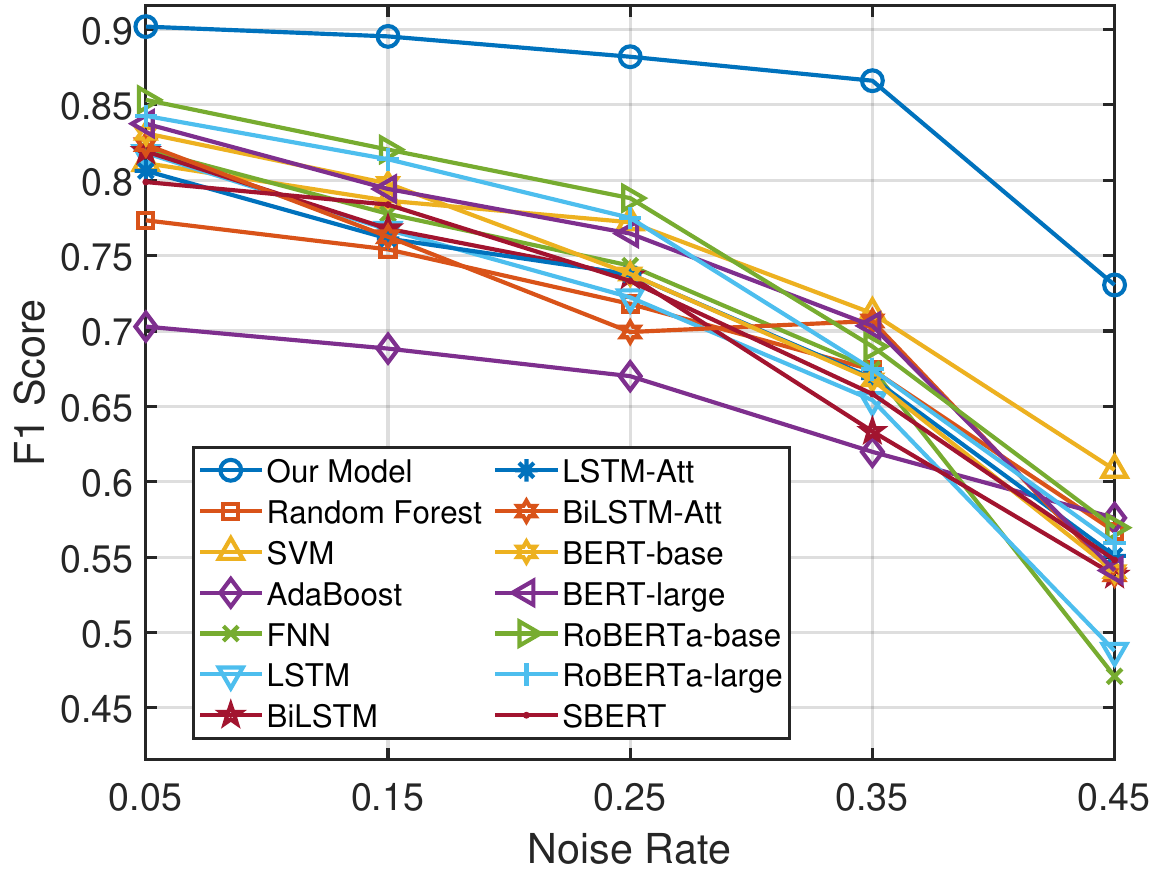}
	\caption{Results of noise experiment}
	\label{fig4}
\end{figure}

The experimental results are shown in Fig.~\ref{fig4}. It is clear that the performance of all models declines after introducing noise, as label errors directly affect the model's ability to capture the true data distribution during learning. Consequently, the models learn incorrect patterns, leading to decreased performance. However, the proposed model shows the smallest decline and consistently outperforms the others, demonstrating its superiority in handling noisy data. This may be due to our model's structure. By incorporating contextual information and user historical behavior as extra feature inputs, our model may better understand the true context behind the target text when faced with noise, resulting in more accurate predictions. The experimental results demonstrate that our model not only achieves SOTA performance on clean data but also maintains strong robustness and stable performance under label-noise conditions, making it well-suited for real-world social media sarcasm detection scenarios where noisy labels are inevitable.

\subsection{Robustness Experiment}

In the first experiment, we used a balanced dataset for model comparison. However, in real-world scenarios, the proportion of sarcastic comments may be much smaller. To verify the robustness of the proposed model, we tested its performance with varying proportions of sarcastic comments. We kept the dataset size fixed at 20,000 and incrementally increased the percentage of sarcastic comments from 10\% to 90\% in 10\% steps.

\begin{figure}[htbp]
	\centering
	\includegraphics[width=2.4in]{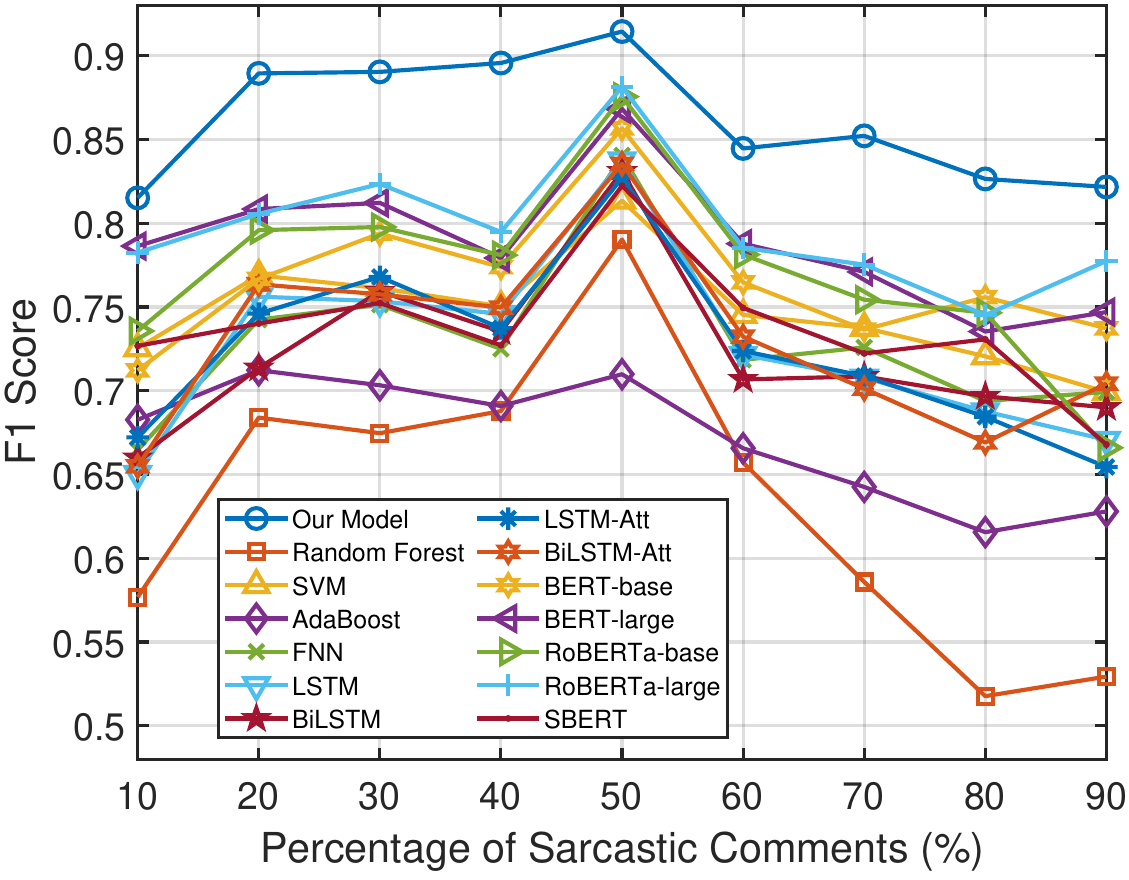}
	\caption{Results of robustness experiment}
	\label{fig3}
\end{figure}

As shown in Fig.~\ref{fig3}, our method and RoBERTa-large perform best when the proportion of sarcastic comments reaches 50\%. Even at relatively high proportions of sarcastic comments, our model consistently outperforms others, demonstrating greater robustness in practical scenarios. This may be due to the multi-dimensional feature fusion in our model, which combines textual content, contextual information, and user historical behavior features. These factors work together to enable the model to more accurately capture the subtle semantics of sarcasm and the user's linguistic patterns, reducing reliance on the dataset's proportion. The RoBERTa-large model performs relatively well. Compared with traditional machine learning approaches that only rely on text features or general deep learning models that introduce limited contexts, RoBERTa-large, through large-scale pre-trained language modeling, can effectively capture the complex semantic implications and contextual dependencies in the text. Meanwhile, by integrating users' historical behaviors with multi-level context information, the model's robustness against intention inconsistencies in satirical expressions has been further enhanced.

\begin{figure}[htbp]
	\centering
	\includegraphics[width=2.4in]{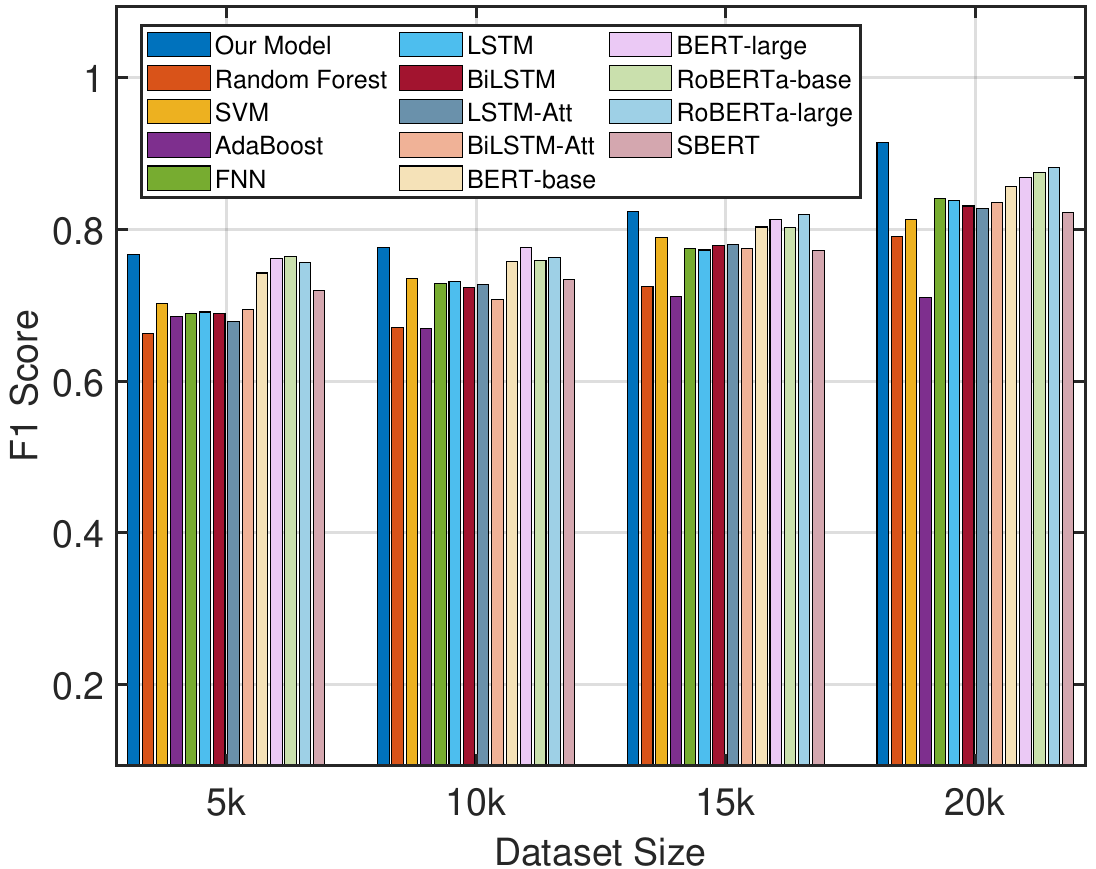}
	\caption{Results of dataset size experiments}
	\label{fig6}
\end{figure}

\subsection{Impact of Dataset Size on Detection Performance}

To validate the effectiveness of the \textit{SinaSarc} dataset generated by our GAN and LLM-driven data augmentation framework for Chinese sarcasm detection, we conducted an experiment to expand the dataset size, from 5,000 to 20,000. Our model was compared with several SOTA models on these datasets. The results are shown in Fig.~\ref{fig6}.

Notably, our proposed model consistently outperforms all baseline methods across different dataset scales, demonstrating strong data efficiency. Even with only 5k training samples, our model achieves a relatively high F1 score, already surpassing most baselines trained on the full 20k dataset. As the dataset size increases to 20k, the model reaches its peak performance while maintaining a clear advantage over competing methods. This superior data efficiency can be attributed to the core design of our approach. Unlike conventional models that rely primarily on text (T) or text combined with context (T+C), our model integrates text, context, and user historical behavior (T+C+U) to capture users’ long-term linguistic patterns. These user-level signals provide personalized and stable cues that complement limited textual information, enabling the model to learn more robust sarcasm representations even under low-resource settings. In contrast, although pre-trained models and LLMs can leverage rich contextual information, they do not explicitly model such user-specific linguistic patterns, and thus consistently underperform our approach across all dataset scales.

\begin{table*}[!t]
	\centering
	\caption{Examples of Sarcasm Detection Results}
	\label{tab5}
	\begin{CJK}{UTF8}{gbsn}
		\begin{tabularx}{\textwidth}{|>{\centering\arraybackslash}m{0.3cm} 
				|>{\centering\arraybackslash}m{1.5cm} 
				|>{\centering\arraybackslash}m{2.7cm} 
				|>{\raggedright\arraybackslash}m{7cm} 
				|>{\centering\arraybackslash}m{2cm} 
				|>{\centering\arraybackslash}m{2cm}|}
			\hline
			\textbf{No.} & \textbf{Topic} & \textbf{Comment Hierarchy} & \centering \textbf{Comments} & \textbf{Correct results} & \textbf{Test results} \\
			\hline
			1 & Lifestyle & Nested comment & 你是真的懂设计 (You \textit{really} know design) & Sarcastic & Non-sarcastic \\
			\hline
			2 & Politics & Nested comment & 成都省的GDP当然高了 (Chengdu province's GDP is certainly high) & Sarcastic & Non-sarcastic \\
			\hline
			3 & Entertainment & Top-level comment & 这张票真值了，两个小时的电影感觉看了五个小时 (This movie ticket was totally worth it—two hours that felt like five!) & Sarcastic & Non-sarcastic \\
			\hline
			4 & Relationships & Nested comment & 男人到某一阶段就会自动解锁历史学家、哲学家、政治家等角色 (Men just \textit{naturally} become historians, philosophers, and statesmen at a certain stage in life.) & Sarcastic & Non-sarcastic \\
			\hline
			5 & Public Incidents & Top-level comment & 这届网友可真``苛刻”，连救援人员的休息时间都要盯着，就怕他们累着 (This group of netizens is really ``harsh" — they even keep an eye on the rescuers' rest time, fearing they might get tired) & Non-sarcastic & Sarcastic \\
			\hline
		\end{tabularx}
	\end{CJK}
\end{table*}

\subsection{Case Analysis}
As shown in Table~\ref{tab5}, to gain a deeper understanding of the complexity of sarcasm detection and explore potential improvements to our proposed method, we analyzed the model's detection results. We randomly selected 1 incorrect sample from each topic to analyze the causes of the errors

In Case 1, the model misclassified a sarcastic nested comment as non-sarcastic. The target comment \begin{CJK}{UTF8}{gbsn}``你是真的懂设计”\end{CJK} (You \textit{really} know design) is sarcastic because it conveys irony through exaggerated praise. The expression appears positive on the surface but, in context, implies criticism of the design. The model failed to capture this implicit reversal of meaning and relied heavily on the literal positive wording, leading to an incorrect judgment.

In Cases 2, the model misclassified the sarcastic comment as non-sarcastic because the target comment contained expressions that required external background knowledge to accurately identify its sarcastic intent. \begin{CJK}{UTF8}{gbsn}``成都省”\end{CJK}(Chengdu Province) is a sarcastic term used for Chengdu city, reflecting public dissatisfaction with the resource allocation favoring it over other cities in Sichuan. 

In Case 3, the sarcastic comment \begin{CJK}{UTF8}{gbsn}``这张票真值了，两个小时的电影感觉看了五个小时”\end{CJK} (This movie ticket was totally worth it—two hours that felt like five!) was also misclassified as non-sarcastic. This statement uses hyperbole to express dissatisfaction with the length and quality of the movie. Although the wording contains no explicit negative terms, the sarcastic intent lies in the contrast between the literal meaning of \begin{CJK}{UTF8}{gbsn}“值”\end{CJK} (worth it) and the actual negative experience of time dragging on. The model failed to recognize this rhetorical exaggeration and instead interpreted the surface-level meaning, which resulted in misclassification.

\begin{figure*}[htpb]
	\centering

	\subfloat[LSTM\label{fig:LSTM}]{%
		\includegraphics[width=0.238\textwidth]{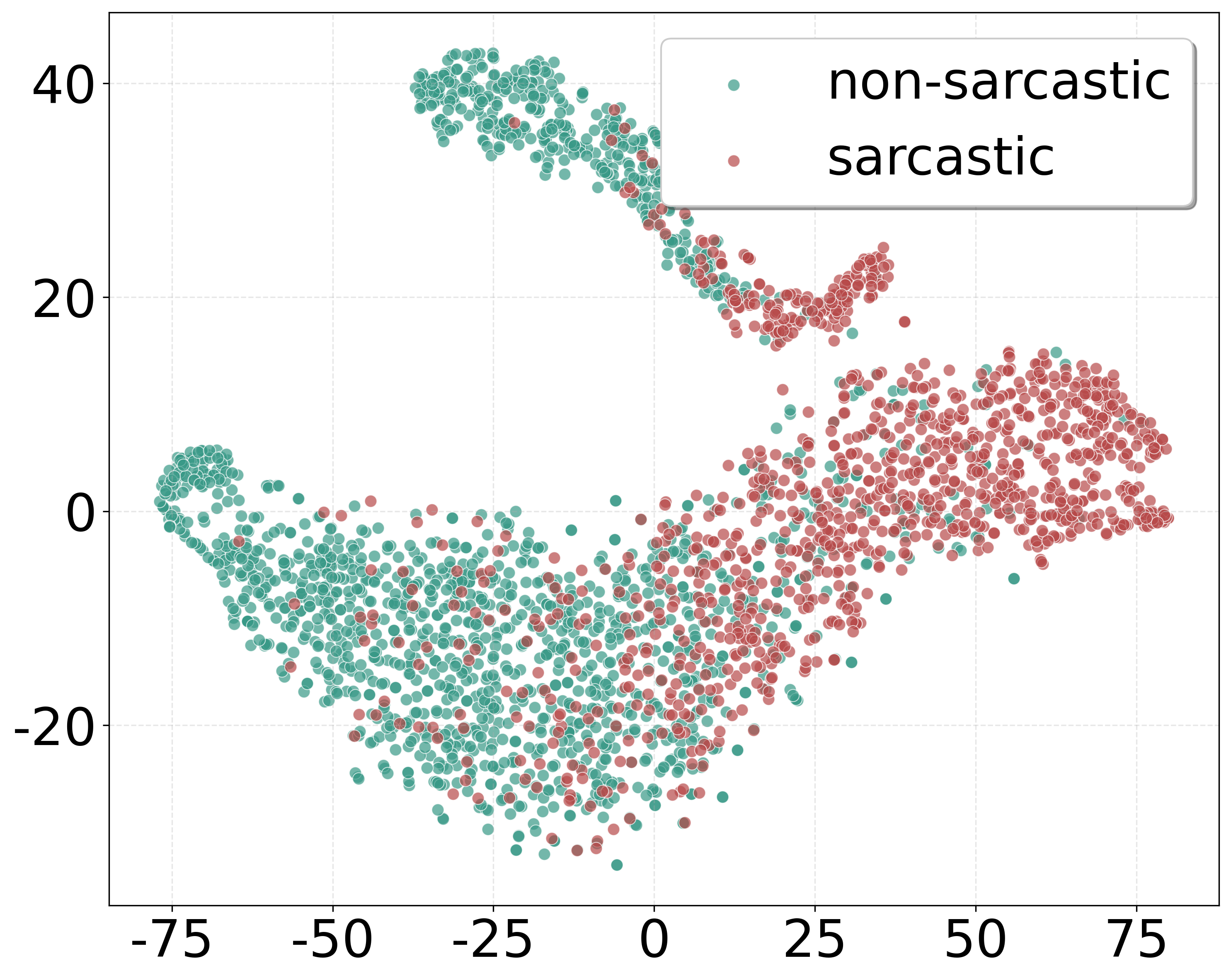}
	}
	\hspace{0.00\textwidth}
	\subfloat[BiLSTM\label{fig:BiLSTM}]{%
		\includegraphics[width=0.238\textwidth]{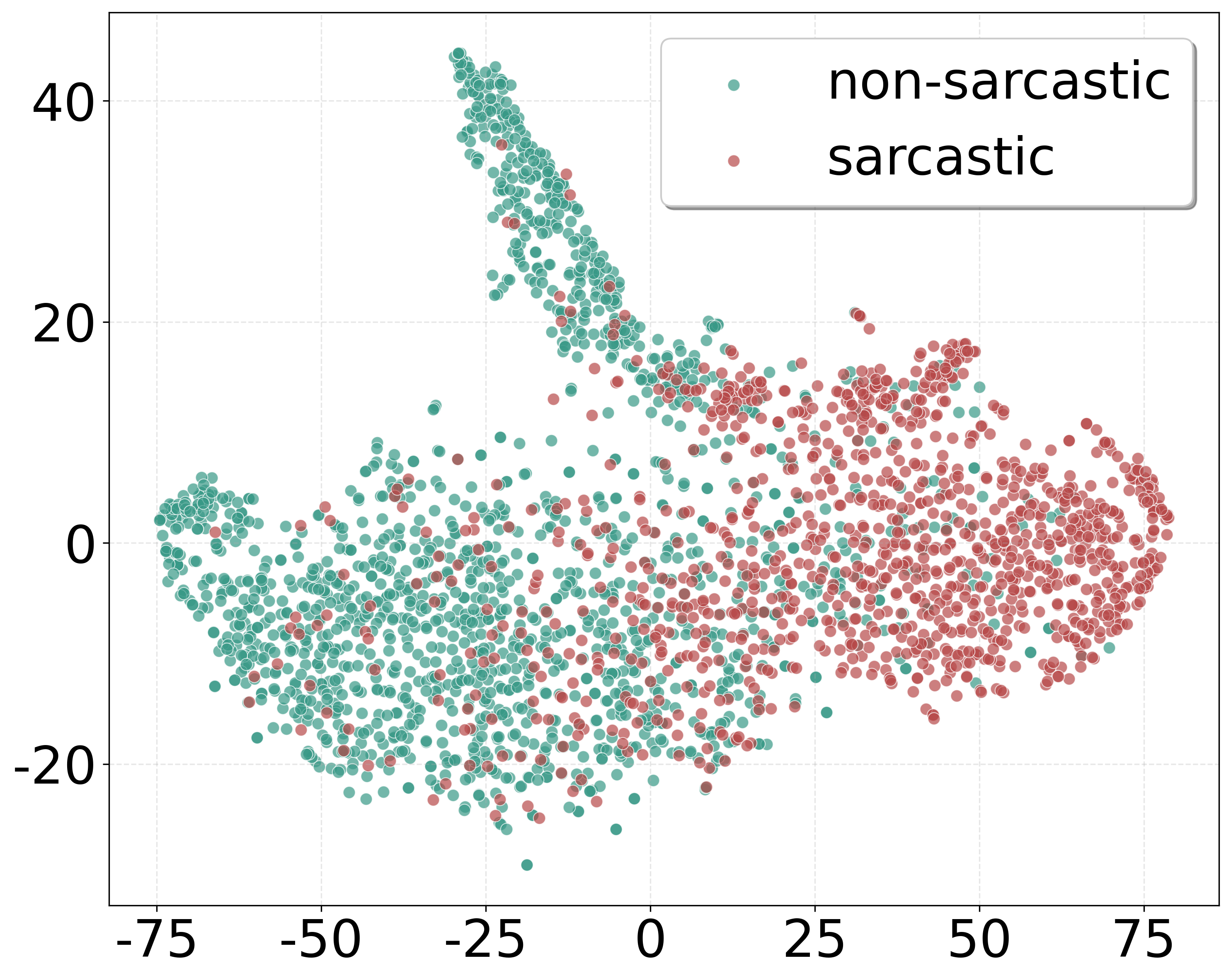}
	}
	\hspace{0.00\textwidth}
	\subfloat[BERT-base\label{fig:BERT-base}]{%
		\includegraphics[width=0.238\textwidth]{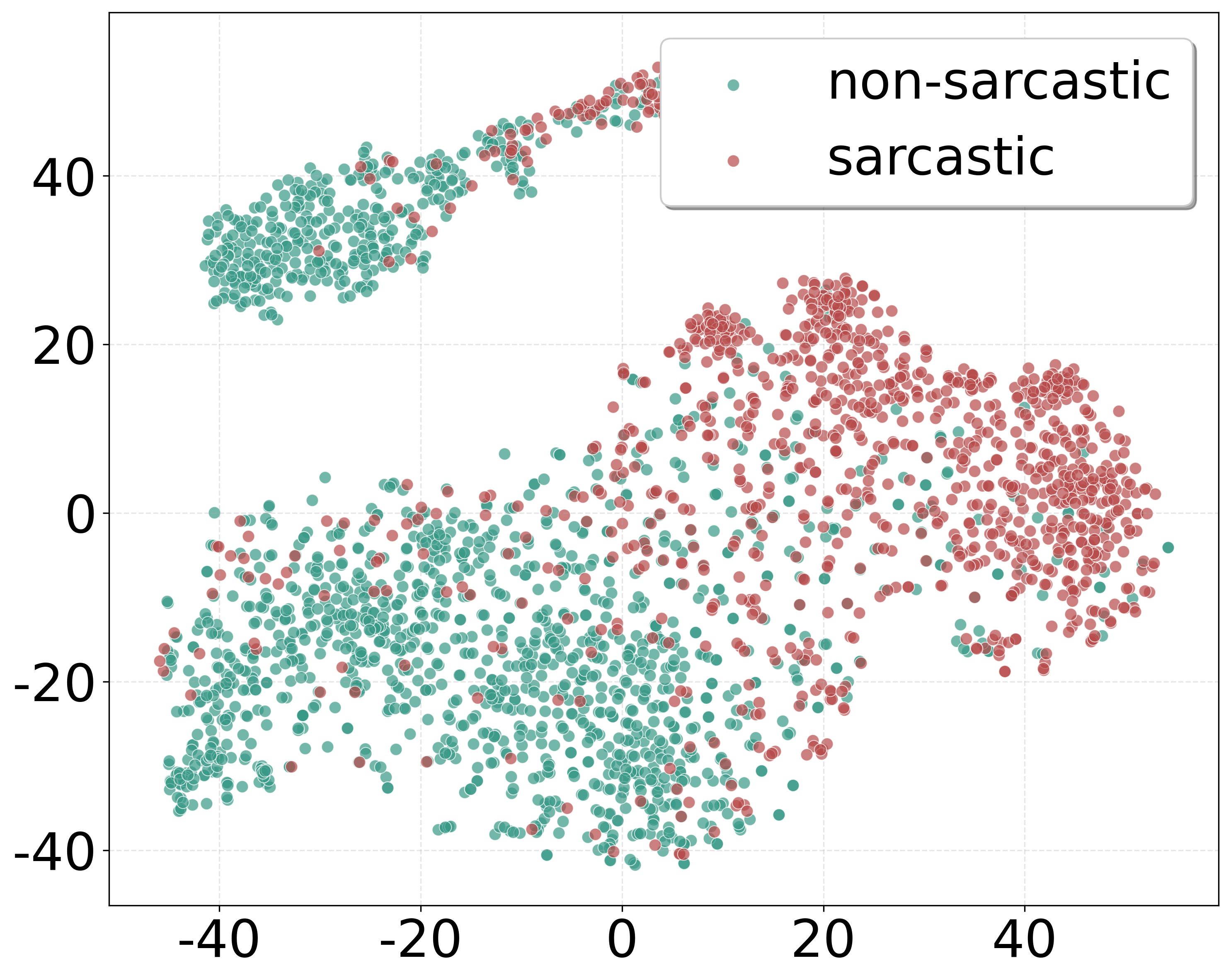}
	}
	\hspace{0.00\textwidth}
	\subfloat[BERT-large\label{fig:BERT-large}]{%
		\includegraphics[width=0.238\textwidth]{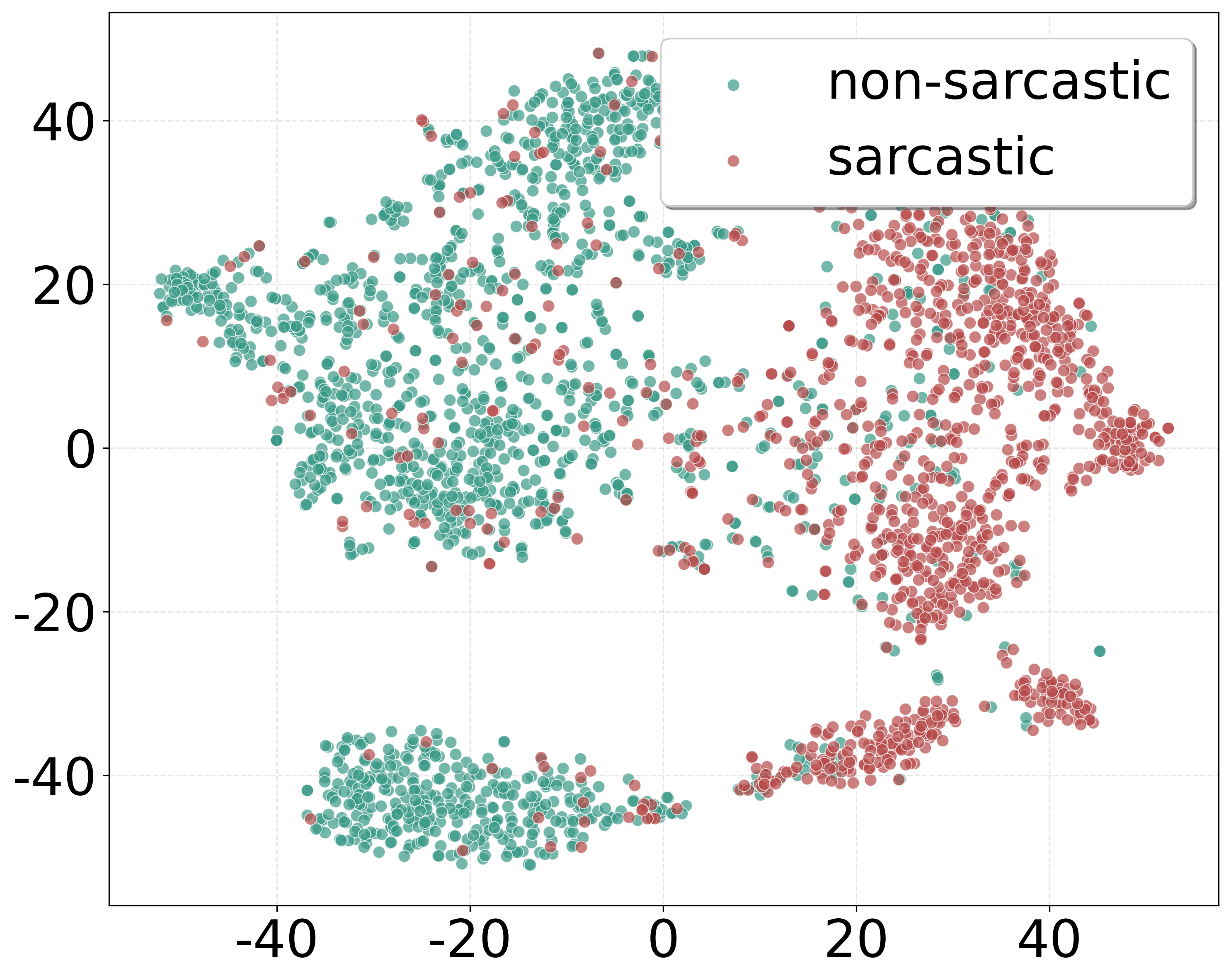}
	}
	
	\vspace{0.00cm}

	\subfloat[RoBERTa-base\label{fig:RoBERTa-base}]{%
		\includegraphics[width=0.238\textwidth]{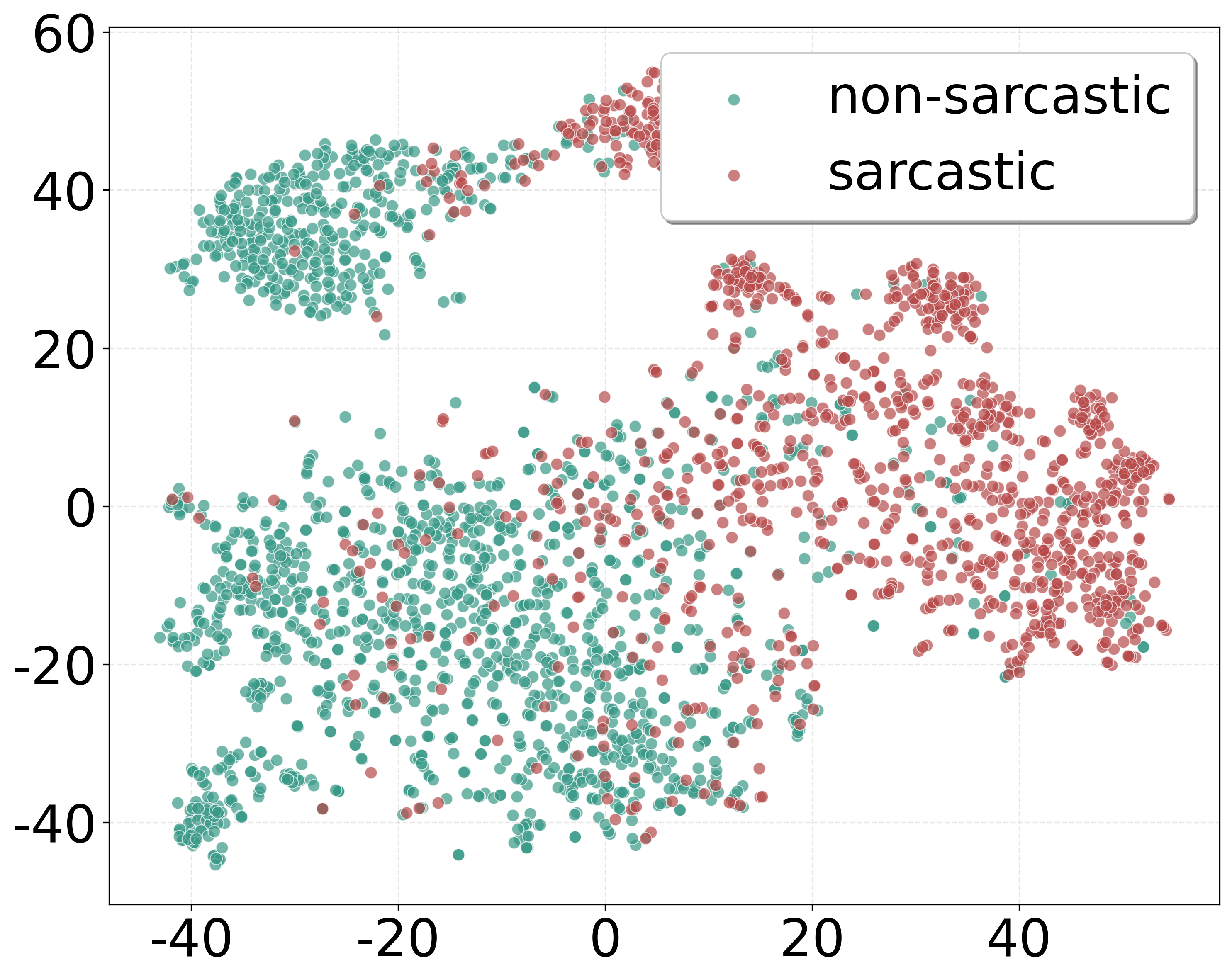}
	}
	\hspace{0.00\textwidth}
	\subfloat[RoBERTa-large\label{fig:RoBERTa-large}]{%
		\includegraphics[width=0.238\textwidth]{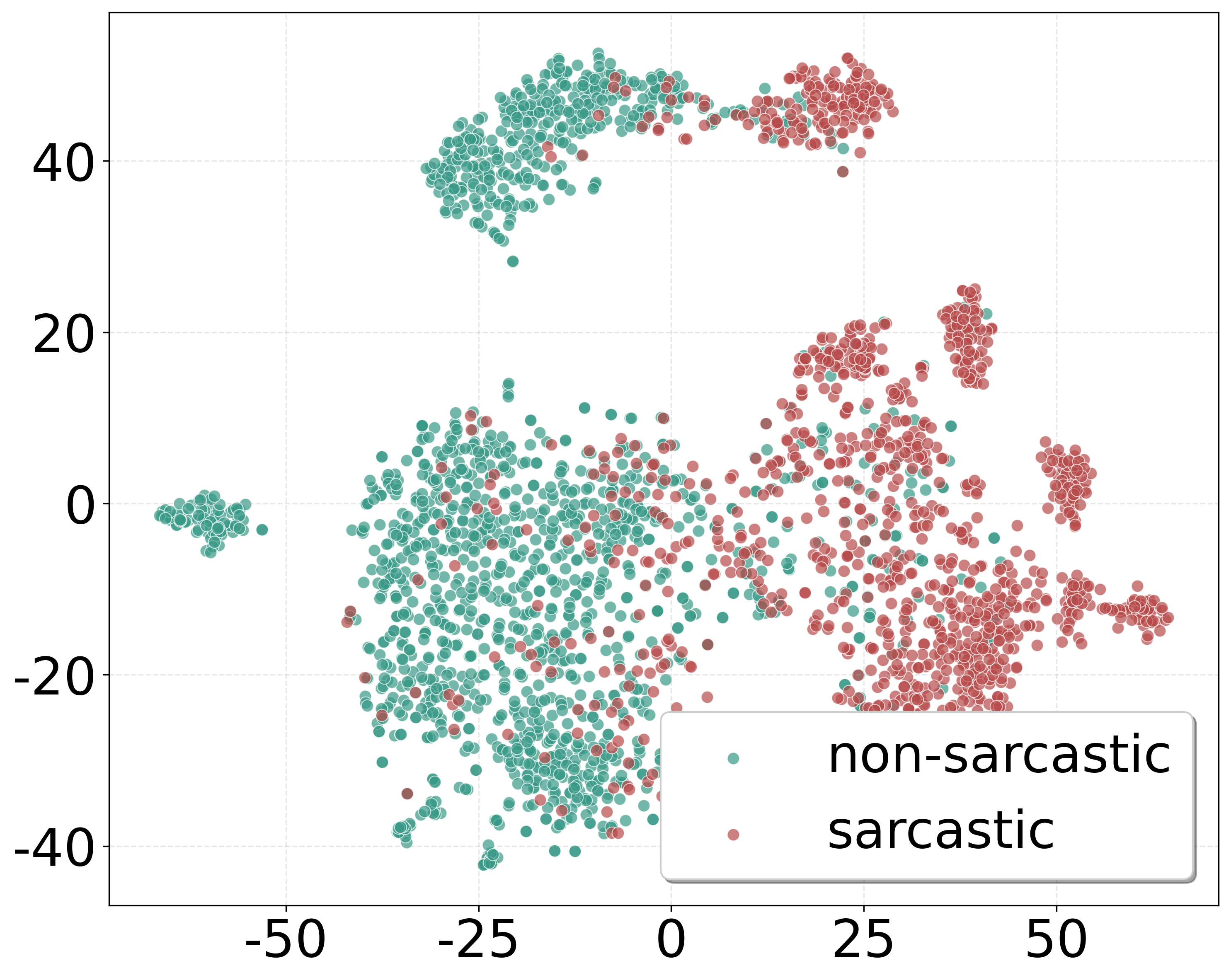}
	}
	\hspace{0.00\textwidth}
	\subfloat[SBERT\label{fig:SBERT}]{%
		\includegraphics[width=0.238\textwidth]{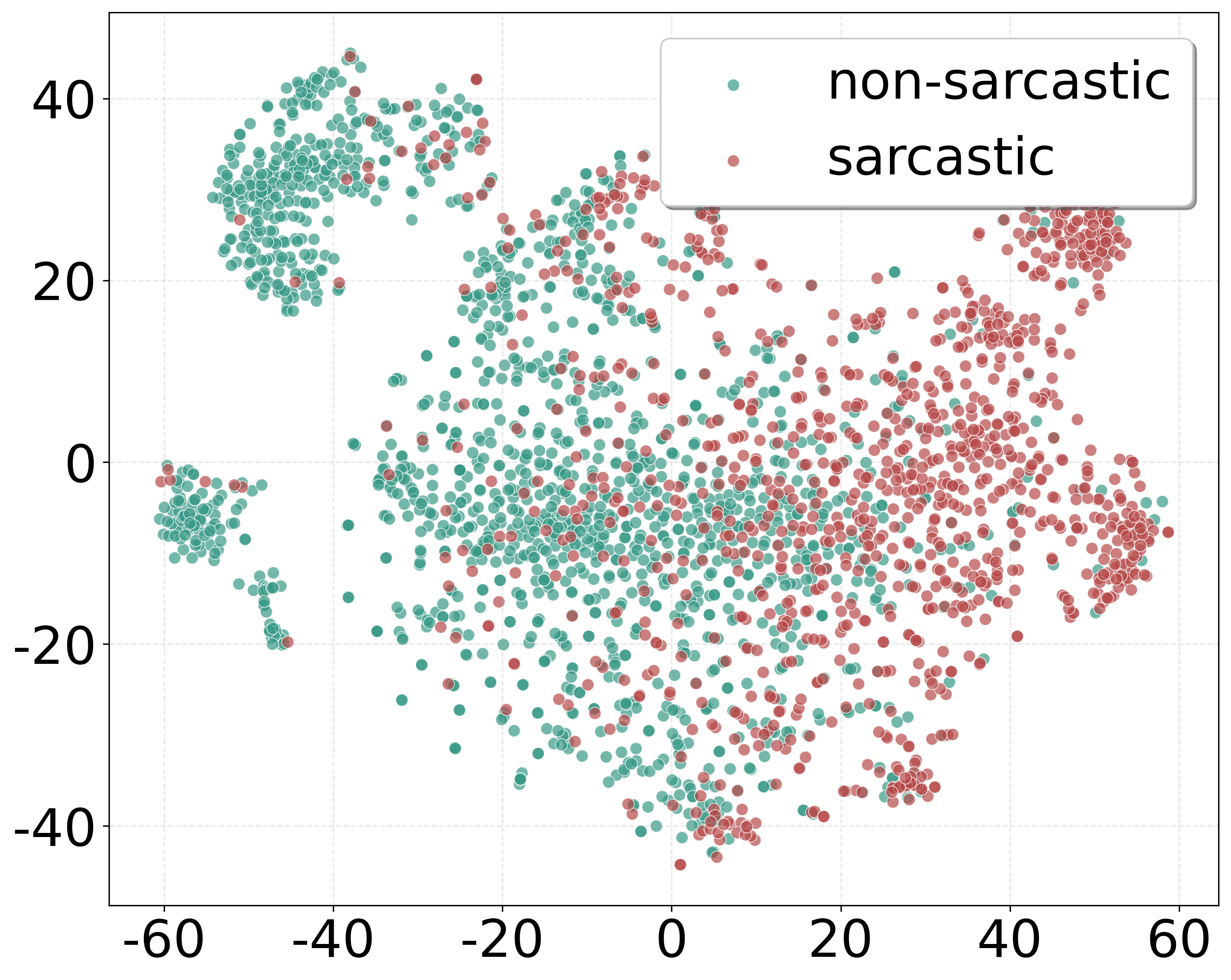}
	}
	\hspace{0.00\textwidth}
	\subfloat[Our Model\label{fig:OurModel}]{% 
		\includegraphics[width=0.238\textwidth]{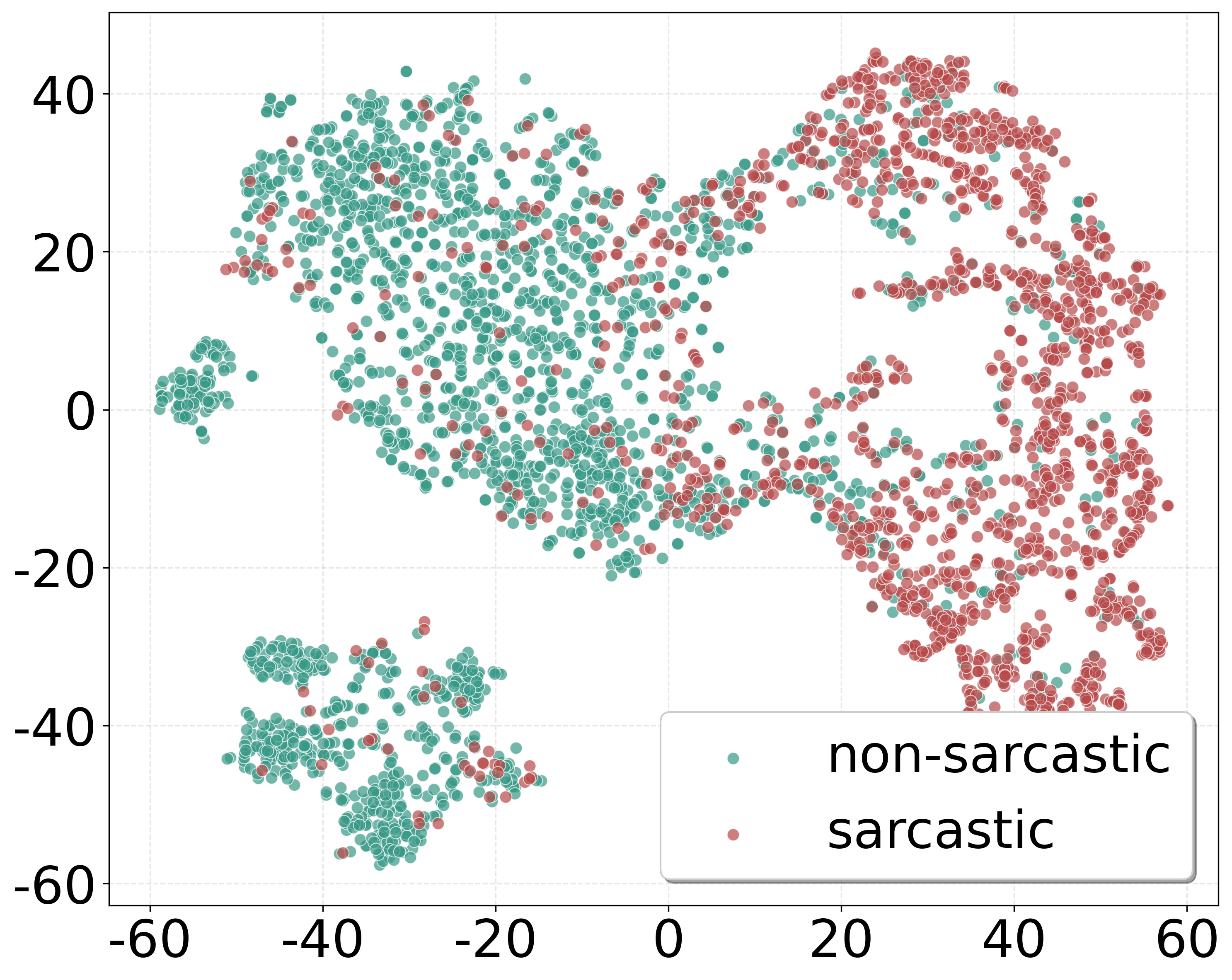}
	}
	
	\caption{Results of characterization learning experiment}
	\label{fig5}
\end{figure*}

Case 4 presented a metaphorical expression that caused the model to misclassify a sarcastic comment as non-sarcastic. This case critiques certain men who express excessively to project a learned image, which can be off-putting. The comment lacked obvious sarcastic indicators and relied on subtle metaphorical language. In Case 5, the model mistakenly classified non-sarcastic comments as sarcastic due to the presence of quotation marks, emotional reversals, and other obvious indicators of sarcasm.

These case analyses reveal that the model has limitations in capturing sarcasm that relies on external background knowledge, rhetorical strategies, and metaphorical expressions. To address these shortcomings, future work will focus on integrating external knowledge graphs to provide richer contextual understanding, developing approaches to model and interpret rhetorical devices, and establishing mechanisms to align metaphors with their underlying referents. 

\subsection{Representation Learning Effect Experiment}
To qualitatively assess the feature representations learned by our model, we conduct a representation learning analysis and visualize the high-dimensional embeddings of sarcastic and non-sarcastic samples using t-SNE, as shown in Fig.~\ref{fig5}. For comparison, we also present the embeddings produced by several representative baselines.

As observed from the visualization, most baseline models fail to clearly separate sarcastic and non-sarcastic samples in the embedding space. The two classes are heavily intermixed, with substantial overlap and no distinct boundary, indicating limited discriminative capability in capturing the implicit and context-dependent nature of sarcasm. In contrast, our model achieves a clear separation between the two classes. As shown in Fig.~\ref{fig5} (h), sarcastic and non-sarcastic samples form well-defined and minimally overlapping clusters, demonstrating more discriminative and structured representations. This improvement can be attributed to the integration of user historical behavior, which enables the model to capture users’ long-term linguistic patterns and uncover subtle semantic distinctions that are difficult to identify using text or context alone.

\section{Conclusion}
This study proposes a GAN and LLM-driven data augmentation framework to dynamically model users’ linguistic patterns in Chinese sarcasm detection, and constructs a novel dataset \textit{SinaSarc}. Experimental results demonstrate that modeling users’ linguistic patterns from their historical behavior significantly enhances the accuracy and robustness of sarcasm detection. Nonetheless, the current dataset is domain-constrained, and generated samples may not fully capture the subtlety of natural sarcastic expressions. Future work will focus on extending the dataset to broader domains and incorporating richer user and multimodal features. In addition, it could also incorporate ideas from Masked Emotion Modeling ~\cite{ref106} to further explore the intrinsic connections between the multidimensional emotional characteristics in satirical expressions.

\balance

\bibliographystyle{IEEEtran}

\bibliography{references}

\end{document}